\documentclass[sigconf]{acmart}
\renewcommand\footnotetextcopyrightpermission[1]{}
\usepackage{booktabs} 
\usepackage{subfigure}

\begin{document}
\title{Residual-Guide Network for Single Image Deraining}

\author{Zhiwen Fan}
\affiliation{\institution{School of Information Science and Engineering, Xiamen University, China.}}
\email{waynefan@stu.xmu.edu.cn}

\author{Huafeng Wu}
\affiliation{\institution{School of Information Science and Engineering, Xiamen University, China.}}
\email{whfeng@stu.xmu.edu.cn}

\author{Xueyang Fu}
\affiliation{\institution{School of Information Science and Engineering, Xiamen University, China.}}
\email{fxy@stu.xmu.edu.cn}

\author{Yue Huang}
\affiliation{\institution{School of Information Science and Engineering, Xiamen University, China.}}
\email{yhuang2010@xmu.edu.cn}

\author{Xinghao Ding}
\affiliation{\institution{School of Information Science and Engineering, Xiamen University, China.}}
\email{dxh@xmu.edu.cn}
%
%
%
%
%
%
%

\begin{abstract}
Single image rain streaks removal is extremely important since rainy images adversely affect many computer vision systems. Deep learning based methods have found great success in image deraining tasks. In this paper, we propose a novel residual-guide feature fusion network, called ResGuideNet, for single image deraining that progressively predicts high-quality reconstruction. Specifically, we propose a cascaded network and adopt residuals generated from shallower blocks to guide deeper blocks. By using this strategy, we can obtain a coarse to fine estimation of negative residual as the blocks go deeper. The outputs of different blocks are merged into the final reconstruction. We adopt recursive convolution to build each block and apply supervision to all intermediate results, which enable our model to achieve promising performance on synthetic and real-world data while using fewer parameters than previous required. ResGuideNet is detachable to meet different rainy conditions. For images with light rain streaks and limited computational resource at test time, we can obtain a decent performance even with several building blocks.  Experiments validate that ResGuideNet can benefit other low- and high-level vision tasks.
\end{abstract}
%
%

\maketitle

\section{Introduction}

Rain streaks degrade visual quality on images and video. Due to the block and blurred effect to
objects in a rainy image, undesirable result of many outdoor computer vision applications
like object detection \cite{Kaiming} will be adversely affected. It is because most existing algorithms are trained
with well-controlled conditions. Thus, designing an effective method for removing rain streaks is desirable for a wide range
of practical applications. Deep learning has been introduced for this problem since Convolutional neural networks (CNN) have
proven powerful for a variety of vision tasks.

However, existing models in rain streaks removal tasks tend to learn negative residual within a single model, these models have to be carefully designed with tones of parameters to capture different modalities of rain streaks. Also most methods optimized with Euclidean distance that will inevitably generate blurry predictions since the per-pixel losses do not close to
\begin{figure}
\subfigure[Rainy image] { {\includegraphics[width=1.58in]{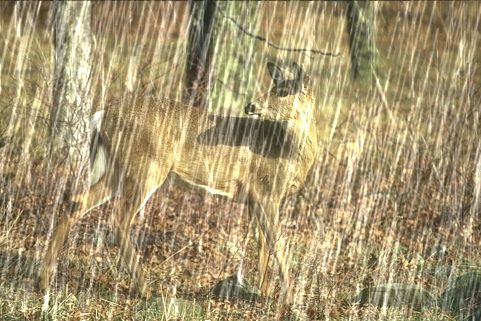}}}
\subfigure[Output of block$_1$] { {\includegraphics[ width=1.58in]{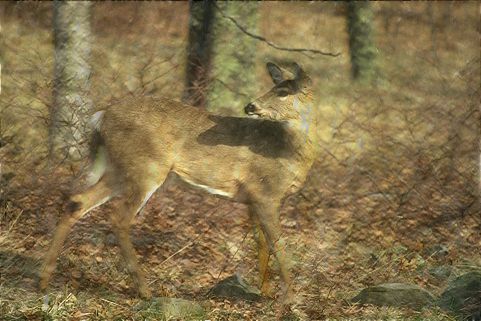}}}
\subfigure[Output of block$_3$] { {\includegraphics[width=1.58in]{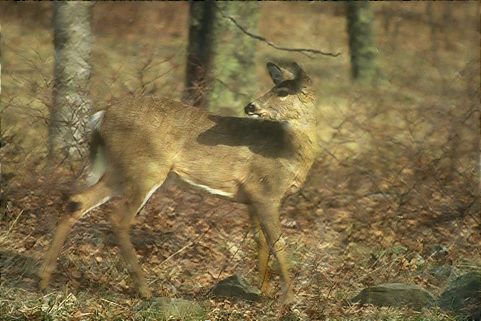}}}
\subfigure[Output of block$_5$] { {\includegraphics[width=1.58in]{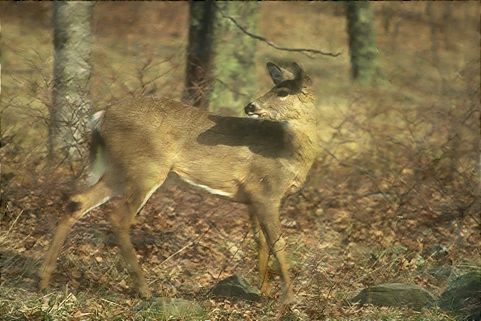}}}
\caption{Progressive high-quality result as blocks go deeper, the SSIM of the output of block$_1$, block$_3$, block$_5$ is 0.927, 0.935, 0.943, respectively.}

\end{figure}
perceptual difference between output and ground-truth images as human visual perception \cite{johnson2016perceptual}.
Further, it is wasteful to utilize a resource-hungry model to meet all kinds of demands for rain streak removal
tasks. For example, under light rainy conditions, a simple model can obtain a
decent derain result, whereas a heavy rainy image should be handled with a
computationally intensive model to detect rain streaks with different shapes
and scales.

To address above drawbacks, we propose the residual-guide feature
fusion network (ResGuideNet) in a cascaded architecture. Each block contains a global shortcut to predict residual \cite{Fu} which can make
the learning process much easier.  However, a simply cascaded basic building blocks is of difficulty
to improve the reconstructed quality in deeper blocks. We conjecture that it is because a cascaded architecture may lost valuable intermediate
reconstruction features which makes the deeper blocks difficult to learn new rain streak pattern.
We then proposed to concatenate the predicted residuals from shallower to the deeper blocks. By using this simple operation, the shallower residuals can guide deeper predictions to generate a finer estimation as shown in Figure 1.

In addition, we apply supervision to all intermediate outputs which can obtain a coarse-to-fine residual as the blocks go deeper. The basic rain streak removal block is
based on recursive computations with a proper shortcut strategy to reduce the number of network parameter while keeping good
derain performance. The final recovered image merges all outputs of intermediate reconstruction which can be viewed as an ensemble learning.
\begin{figure*}
\begin{center}
\subfigure { {\includegraphics[width=1\textwidth]{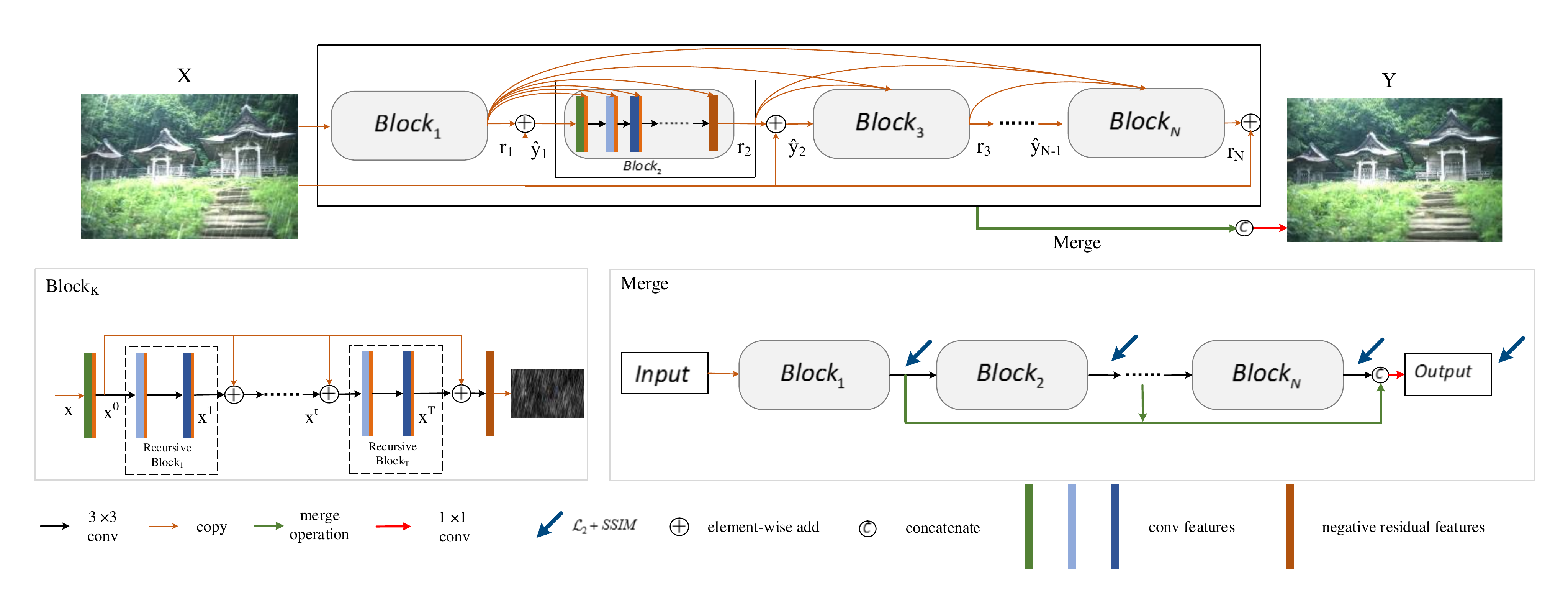}}}
\caption{The proposed structure of our rain streak residual-guide network(ResGuideNet)}
\end{center}
\end{figure*}

The contributions of our paper are three-fold:
\begin{enumerate}
 \item We build a single and separable network that can handle different rainy conditions. By maintaining negative residual
 features in shallow blocks to deep blocks, a coarse-to-fine estimation of negative rain streaks residual can be obtained.
  As the application scenario changes, the user can detach a portion of our model to meet varying computational requirements at test.
 \item We apply supervision to all the intermediate and final reconstructions with a combined loss function.
The model combines all intermediate results to obtain the final result, which can be viewed as ensemble learning.
 \item We discuss how ResGuideNet can be applied to other low-level
vision tasks including denoising and the reconstructed images could
benefit down-stream applications such as object detection.
\end{enumerate}

\section{Related Works}
Depending on the input format, existing rain streak removal algorithm can be roughly categorized into video-based
methods and single-image methods. For video-based methods \cite{Barnum} \cite{Bossu}
 \cite{J} \cite{Garg} \cite{Santhaseelan},
inter-frame information between adjacent frames is leveraged
to identify rainy region and remove rain streaks.

Removing rain streaks from single-image is more challenging since less
information can be utilized. \cite{Kang} attempt to extract rain streaks and background
details from high-frequency layer by sparse-coding based dictionary learning.
\cite{Luo} proposed a framework to rain removal based on discriminative sparse coding.
\cite{Li} learn background from pre-collected natural images and rains from rainy images
by utilizing two Gaussian mixture models (GMMs).

Deep learning has also introduced for restoration problems and convolutional neural networks (CNN) have
found great success in processing many computer vision problems. The first CNN-based method for single
image deraining was introduced by \cite{fu2017clearing}. The authors build a relative shallow network with
3 layers to learn the mapping function. In \cite{Fu}, combining with ResNet \cite{24} \cite{25}, the authors present a deep
detail network(DDN) to learn residual with the high frequency part of rainy images. In \cite{ID-CGAN}, the author proposed a conditional GAN-based
algorithm for removal of rain streak from a single image. \cite{joder}learn binary rain region mask rand remove the rain streaks simultaneously through a multi-scale network(JORDER). \cite{DID-MDN} utilize the rain density information with a multi-stream densely connected network (DID-MDN) for jointly rain-density estimation and deraining. Further, single image
dehaze \cite{dehazenetTIP} \cite{dehazeECCV} \cite{adoCVPR} \cite{dehaze_hezhang} achieving promising result by introducing deep learning models.

In image restoration field, achieving good performance with a moderate number of network parameters is an important goal for designing
a deep neural network, \cite{19} \cite{20} \cite{21}proposed to reuse the same convolutional filter weight to
learn hierarchical feature representation. In order to avoid gradient vanishing problems and reduce the total
parameters for very large deep models, \cite{22} \cite{23} proposed to use recursive computation with proper supervision and shortcut to achieve state-of-the-art
performance in single-image super resolution while using few parameters.

\section{Method}

\subsection{Motivation}
Since rain streaks are always overlapped with background texture, most methods tend to learn the negative residual of its input with a complex or carefully designed model.
However, this may lead to an over-smoothed result and need tons of parameters to optimize. Also, it is infeasible to apply a resource-hunger model to process video frame-by-frame for its time-consuming processing. On the other hand, to meet different kind of demands in practical applications, a light weight or detachable network is desirable since their huge number of parameters will limit their application in mobile device, automatic driving and video survillence. However, existing methods use a fixed computational budget to handle both "easy" and "hard" application scenarios. This is less flexible for a model to implement in real-world application.

\begin{figure*}
\subfigure[SSIM:1 ] { {\includegraphics[height=0.56in, width=0.92in]{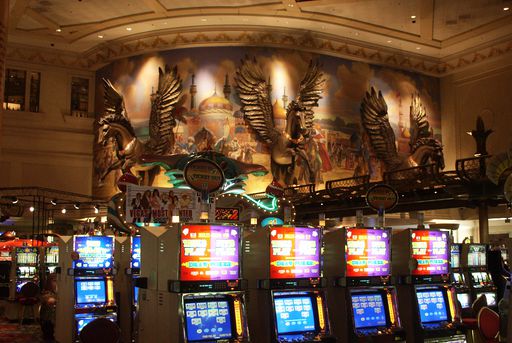}}}
\subfigure[SSIM:0.735] { {\includegraphics[height=0.56in, width=0.92in]{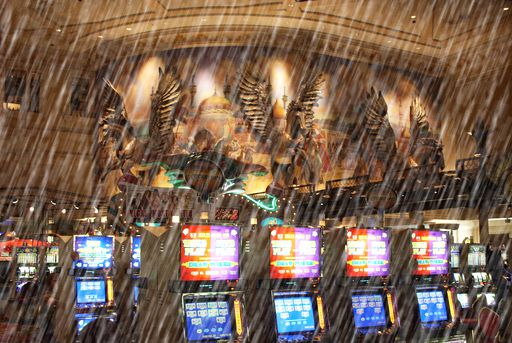}}}
\subfigure[SSIM:0.918] { {\includegraphics[height=0.56in, width=0.92in]{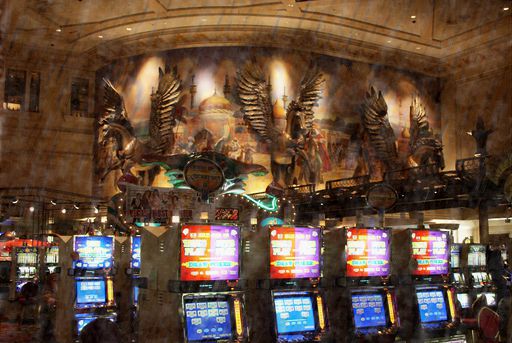}}}
\subfigure[SSIM:0.937] { {\includegraphics[height=0.56in, width=0.92in]{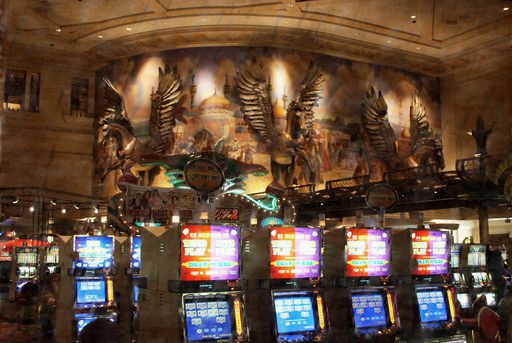}}}
\subfigure[SSIM:0.951] { {\includegraphics[height=0.56in, width=0.92in]{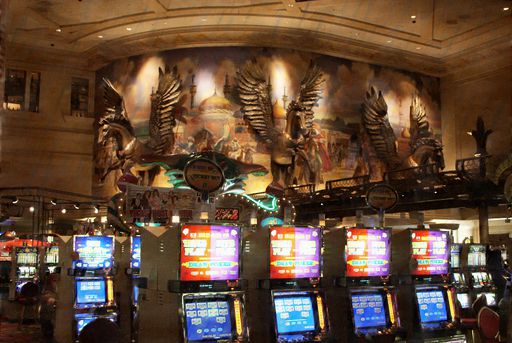}}}
\subfigure[SSIM:0.955] { {\includegraphics[height=0.56in, width=0.92in]{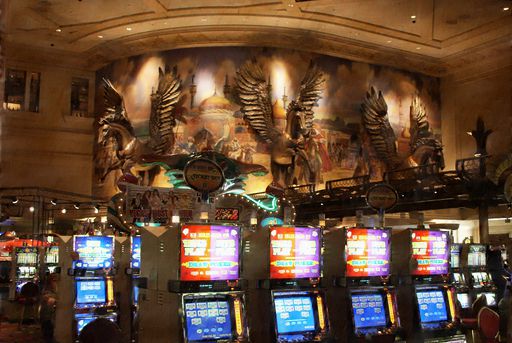}}}
\subfigure[SSIM:0.958] { {\includegraphics[height=0.56in, width=0.92in]{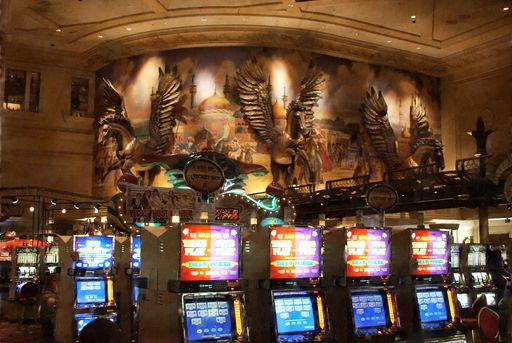}}}

\subfigure[SSIM:1] { {\includegraphics[height=0.56in, width=0.92in]{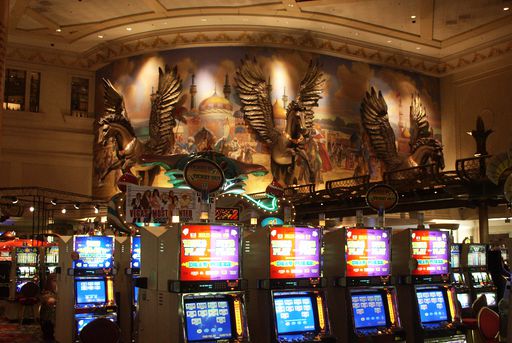}}}
\subfigure[SSIM:0.885] { {\includegraphics[height=0.56in, width=0.92in]{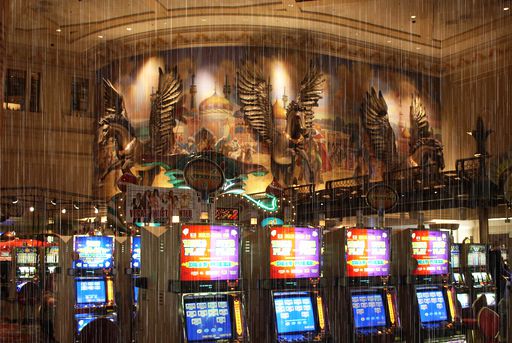}}}
\subfigure[SSIM:0.968] { {\includegraphics[height=0.56in, width=0.92in]{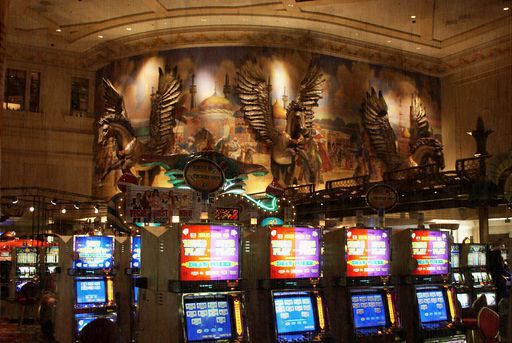}}}
\subfigure[SSIM:0.971] { {\includegraphics[height=0.56in, width=0.92in]{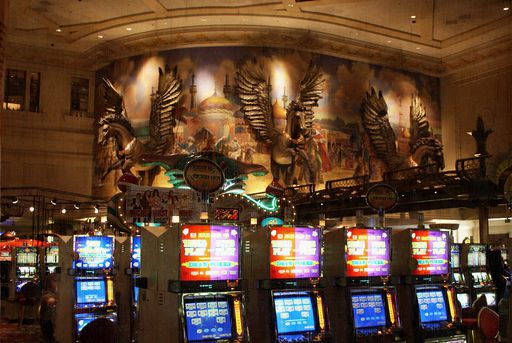}}}
\subfigure[SSIM:0.973] { {\includegraphics[height=0.56in, width=0.92in]{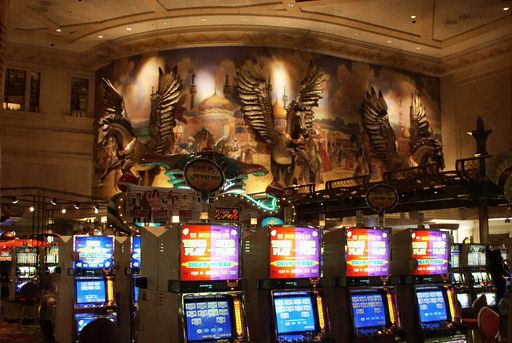}}}
\subfigure[SSIM:0.975] { {\includegraphics[height=0.56in, width=0.92in]{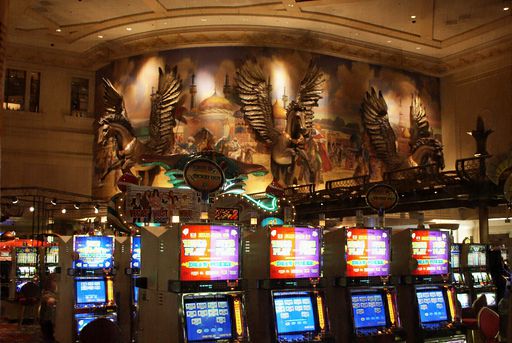}}}
\subfigure[SSIM:0.976] { {\includegraphics[height=0.56in, width=0.92in]{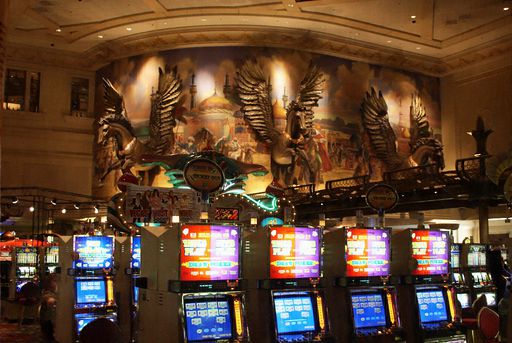}}}
\caption{Comparasion under heavy and light rainy conditions using our ResGuideNet, (a)  and (h) are clean image. (b)and (i) is the synthetic rainy image under heavy rain condition and light rain condition, respectively. (c)-(g) (j)-(n) are the results from block$_1$ to block$_5$ of ResGuideNet under heavy and light rainy conditions, respectively.}
\end{figure*}
%
\begin{figure}
\subfigure { {\includegraphics[height=2in, width=3.2in]{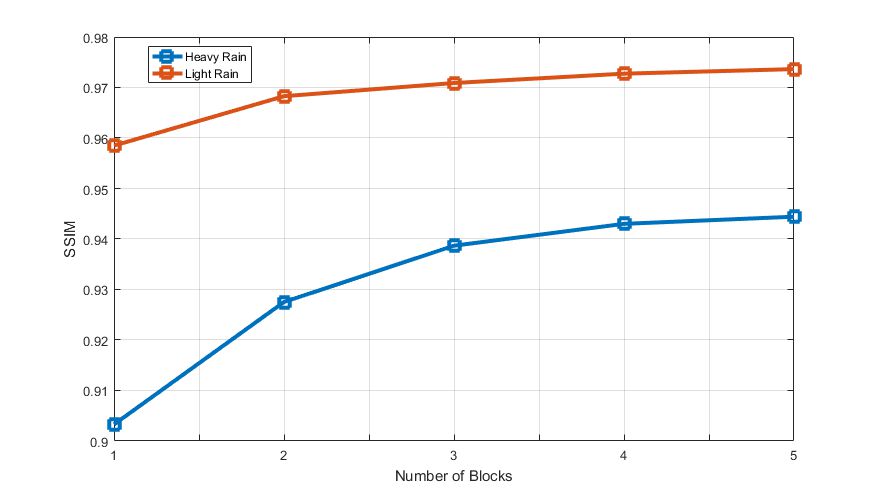}}}
\caption{Deraining result with ResGuideNet under heavy and light rain conditions, we obtain the result on the whole test datasets and averaged them. We can observe the reconstruction does not improve much for light rain condition since block$_2$ to block$_5$ .}
\end{figure}

As is evident in Figure 3, we test our ResGuideNet under heavy and light rain streaks conditions. We can observe our method has a progessively better reconstruction as blocks go deeper. However under light rainy condition, the SSIM \cite{ssimloss} does not improve much since block$_2$ to block$_5$, we can see the averaged SSIM of heavy and light test dataset in Figure 4.

Thus, we would like to build a model that receive good results on all devices, with varying computational constraints of all devices. Furthermore, users can improve the average reconstruction quality by reducing the amount of computation that spent on light rain condition to save up computation for heavy cases.

Motivated by the prior work that has a resouce-efficient implementation \cite{multiscale-densenet}, we aim to construct CNNs that is able to slice the network to meet the computational limitation to process rain streaks under different rainy conditions. Unfortunately, deep neural network is inherently related with the early-existed features. Thus, we build a
model that incorporates a series of deraining sub-networks and progressively generate a cleaner estimation given a rainy input. We can also use a portion of the whole model to handle different rainy conditions.

\subsection{Residual Feature Reuse}

\begin{figure*}
\subfigure[SSIM: 0.716] { {\includegraphics[height=0.67in, width=1.07in]{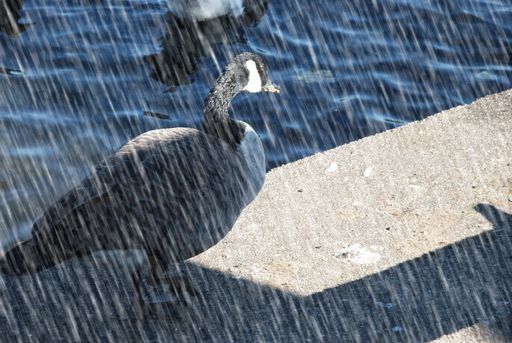}}}
\subfigure[SSIM:0.897] { {\includegraphics[height=0.67in, width=1.07in]{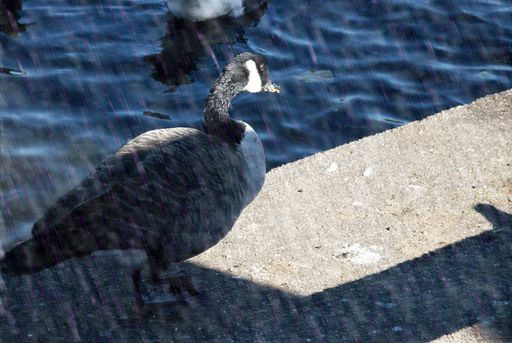}}}
\subfigure[SSIM:0.908] { {\includegraphics[height=0.67in, width=1.07in]{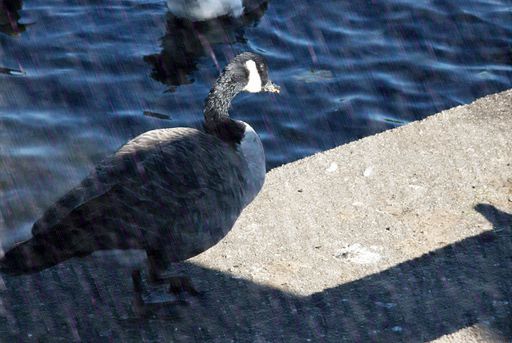}}}
\subfigure[SSIM:0.910] { {\includegraphics[height=0.67in, width=1.07in]{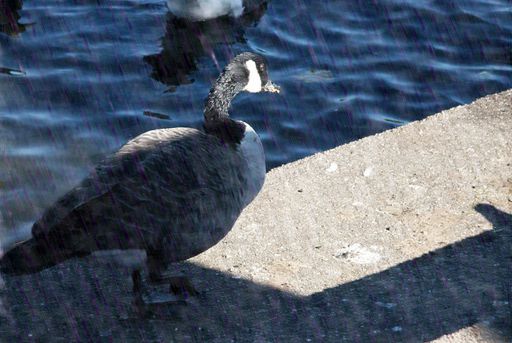}}}
\subfigure[SSIM:0.908] { {\includegraphics[height=0.67in, width=1.07in]{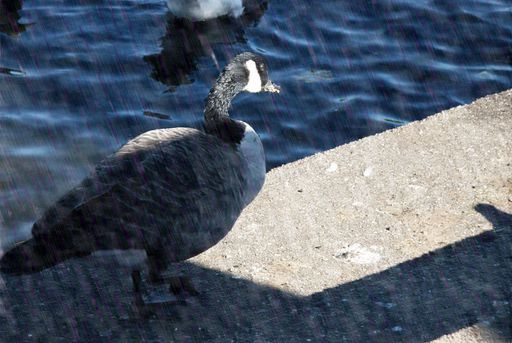}}}
\subfigure[SSIM:0.929] { {\includegraphics[height=0.67in, width=1.07in]{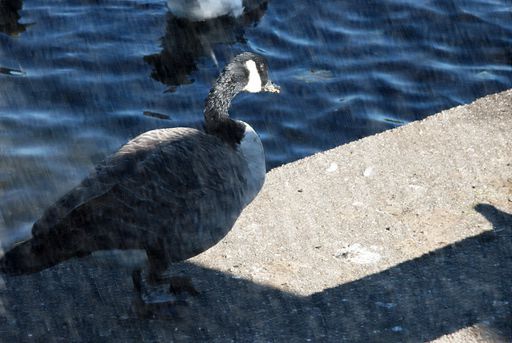}}}
\subfigure[SSIM:1] { {\includegraphics[height=0.67in, width=1.07in]{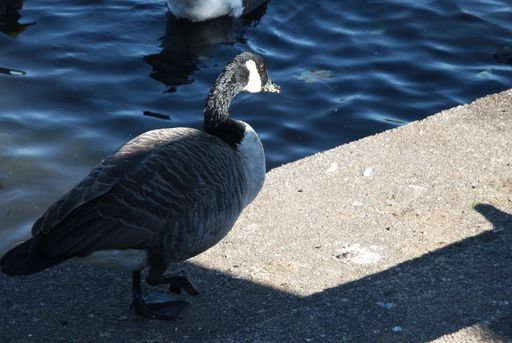}}}
\subfigure[SSIM:0.894] { {\includegraphics[height=0.67in, width=1.07in]{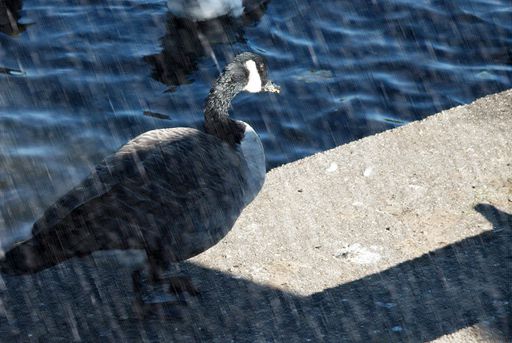}}}
\subfigure[SSIM:0.924] { {\includegraphics[height=0.67in, width=1.07in]{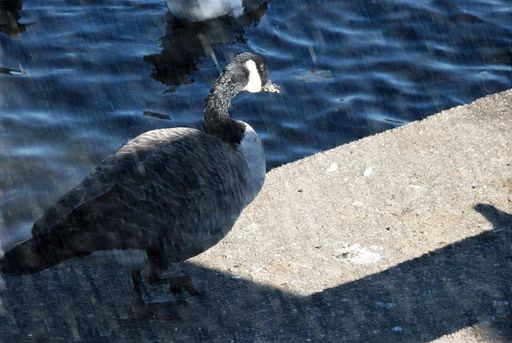}}}
\subfigure[SSIM:0.937] { {\includegraphics[height=0.67in, width=1.07in]{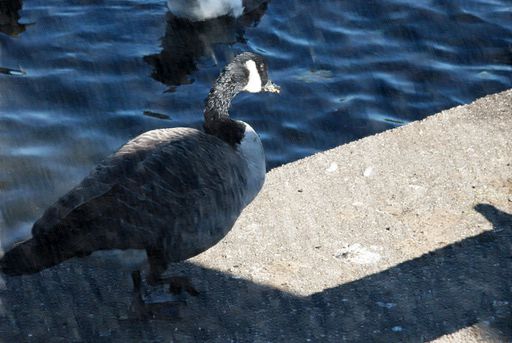}}}
\subfigure[SSIM:0.945] { {\includegraphics[height=0.67in, width=1.07in]{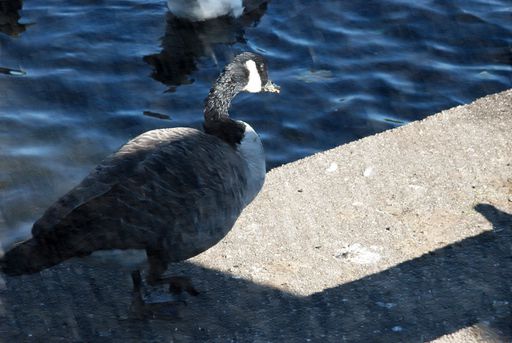}}}
\subfigure[SSIM:0.948] { {\includegraphics[height=0.67in, width=1.07in]{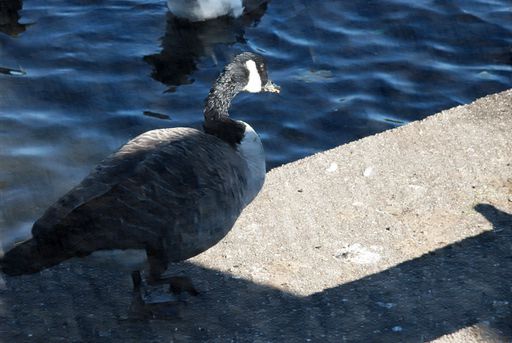}}}
\caption{Comparasion between Baseline model and Baseline model with residual reuse(Baseline-RR) on a single test image, (a) is the synthetic rainy image , (g) is clean image , (b)-(f) are the results from block$_1$ to block$_5$ of Baseline, (h)-(l) are the result from block$_1$ to block$_5$ of Baseline-RR.}
\end{figure*}

A major challenge for deep learning models is its optimization. To address the gradient vanish problem in back propagation, shortcuts have been proposed to stabilize the gradient flow in
deep residual networks (ResNet). By assuming that the residual mapping
is much easier to learn than the original unreferenced mapping,
residual network explicitly learns a residual mapping for a few
stacked layers. With such strategy, deep neural networks can be
easily trained and therefor, ResNet has achieved very impressive performance
on the a number of tasks.
Also, \cite{densenet} proposed to concatenate feature maps densly from lower to deeper layers which can
alleviate the gradient vanishing problem and reduce the number of model parameters. It may be interpreted as there is no
need to relearn redundant features. \cite{memnet} has introduced dense connection in regression tasks and has shown densely connections could benefit the long-term memories and the restoration of mid/high frequency information.

In this paper, we adopt global residual learning with a long shortcut in each block to ease the learning process. Each block consists of several convolutional layers using Leaky
Rectified Linear Units, we refer this architecture as Baseline model. However, simply cascaded blocks cannot obtain promising results. We conjecture that deeper blocks is difficult
 to extract new rain streak patterns and the intermediate reconstructions from lower blocks contain valuable information have lost. To deal with this problem, we suggest to integrate information from previous blocks to deeper ones, to compensate information and further enhance high-frequency signals.


\begin{figure}
\subfigure { {\includegraphics[height=2in, width=3.2in]{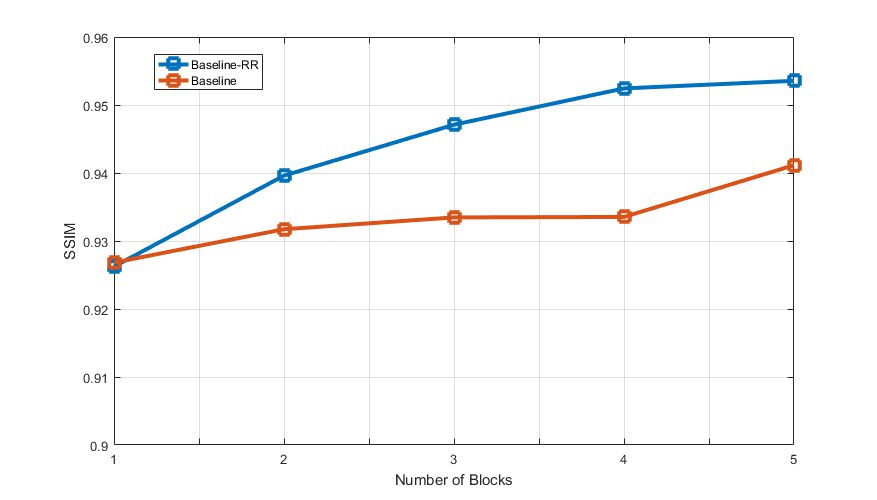}}}
\caption{Comparison of SSIM between Baseline model and Baseline with residual reuse(Baseline-RR).}
\end{figure}

We evaluate the benefit of transitioning from natively cascading deraining blocks(Baseline) to our adopted negative residual reuse(Baseline-RR) by feature fusion strategy using 5 blocks. For fair comparison, we increase the number of feature maps in each building block of Baseline model to have the same parameters with Baseline-RR. We conduct the experiments on the dataset provided by \cite{Fu}. As is clear from the visual quality of reconstruction in Figure 5, Baseline-RR obtain a more eye-pleasing reconstruction and a higher SSIM value as the blocks go deeper. In Figure 6, Baseline-RR obtains a gradual incresement on SSIM as the block becomes deeper, whereas the Baseline model does not possess this property, the SSIM value is based on averging all test images.


\subsection{Loss Function}
Since rain streaks are blend with object edges and background scene, it is hard to distinguish between
rain streaks and objects' structure by simply optimize $\ell_2$ loss function. Per-pixel losses cannot capture perceptual difference
between output and ground-truth images as human visual perception. A model with $\ell_2$ loss tend to result in a blurred reconstruction.
\begin{figure}
\subfigure { {\includegraphics[height=2in, width=3.2in]{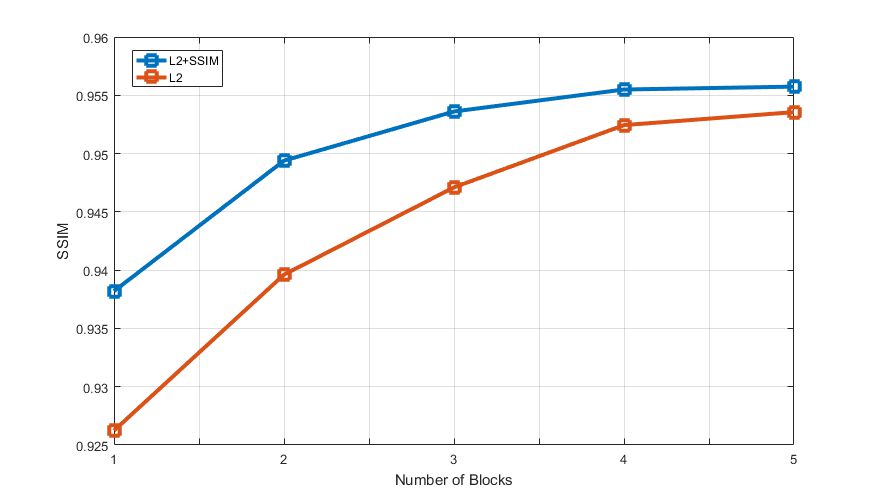}}}
\caption{Comparison of Baseline-RR using different loss function. }
\end{figure}

Therefor, for each block we adopt $\ell_2$+SSIM loss \cite{ssimloss} which can preserve global structure better as well as keeping per-pixel similarity. We minimize the
combination of those loss functions in training stage.
Figure 7 show the effectiveness of the implementation of SSIM loss with $\ell_2$ loss and prove that the supervision to intermediate outputs
could benefit the whole model. Note that, the above experimental result is obtained by averaging 100 test images of dataset \cite{Fu}.

The overall loss function for block$_{k}$ is
\begin{equation}
\begin{aligned}
&L_{MSE_{k}} = \frac{1}{N}\sum_{i=1}^{N}(\|f_{k}(X_{i}, W, b) - Y\|_2^{2}\\
&L_{SSIM_{k}} = \log(1.0 / g(f_{k}(X_{i}, W, b) ,Y) + 1e^{-4})\\
&L_{B_{k}} = L_{MSE_{k}}+\lambda*L_{SSIM_{k}}
\end{aligned}
\end{equation}

Where $N$ is the number of training rainy patches, $k$ indicate the index of block. $X$, $Y$ and $X^i$ indicate rainy patches, corresponding clean patches and the input of block$_i$, respectively. $W$ and $b$ are the parameters in our model that need to tune. $f$ denotes function mapping of each block. $g$ denotes the function of SSIM. $\lambda$ is the hyperparameter that balance the MSE loss and SSIM loss, we set $\lambda$ as 1 via cross-validation that achieving satisfying result.

Note the overall ResGuideNet loss that containing $M+1$ loss function terms if the ResGuideNet contains $M$ blocks
\begin{equation}
\begin{aligned}
&L = \frac{1}{M+1}(\sum_{i=1}^{M}L_{B_{k}}+ L_{Merge} )\\
\end{aligned}
\end{equation}
where the second term in the right side of the equation is the final reconstruction merged by all previous intermediate outputs, with the same format of $L_{B_k}$.
\subsection{Recursive Computation}
As we mentioned above, the trade-off between the number
of parameters and the model performance can be overcame using recursive strategy where
the the nonlinear mapping operator is shared within each block. We adopt two convolutional
operation in each recursive unit. We can write the structure of the input
and output relationship in the $t^{th}$ and $(t+1)^{th}$ recursion ($1 \le t < T$) within each block as
\begin{equation}\label{eq2}
\rm{}\quad{x^t} = g\left( {{x^{t - 1}}} \right),\quad
{x^{t + 1}} = g\left( {{x^t}} \right).
\end{equation}
where $g$ indicates each recursive unit within one block.

However, as the recursions continue, the network depth increases, which introduces a severe
gradient vanish problem that makes training difficult. To solve the gradient vanish problem
as the recursion continues and to propagate information more easily, the output feature map of first
feature extraction Conv+LReLU structure is fed into all subsequent outputs of recursive blocks. We can reformulate the structure as
\begin{equation}\label{eq4}
\rm{}\quad{{x^t} = g\left( {{x^{t - 1}}} \right) + {x^{0}}}, \quad
{{x^{t + 1}} = g\left( {{x^t}} \right) + {x^{0}}}.
\end{equation} the recusive computation is shown in bottom-left of Figure 2.
We evaluate the benefit of transitioning from ResGuideNet without recursion
(ResGuideNet-NRecur) to our adopted ResGuideNet using 5 recursions in each block. We show
the quanlitative result in Figure 8. As is eveident, kernel reuse and propagate all information forward directly from output of the first layer
within each block benefit the restoration process of image content.
\begin{table} [htbp]
 \caption{\label{table1} Performance of ResGuideNet$_5$ and ResGuideNet.\\}
 \centering
 \begin{tabular}{lccl}
  \toprule
     & ResGuideNet$_5$ & ResGuideNet  \\
  \midrule
  SSIM & 0.960 &  0.961   \\
  PSNR & 29.92 &  30.11   \\
  \bottomrule
 \end{tabular}
\end{table}
\begin{figure}
\subfigure[SSIM:0.8181] { {\includegraphics[height=0.9in, width=1.6in]{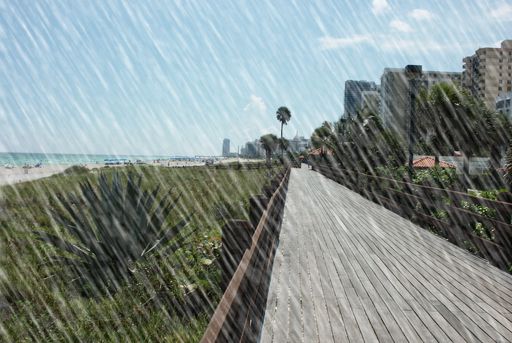}}}
\subfigure[SSIM:0.9506] { {\includegraphics[height=0.9in, width=1.6in]{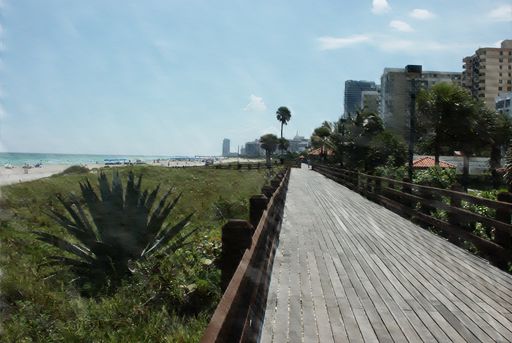}}}
\subfigure[SSIM:1] { {\includegraphics[height=0.9in, width=1.6in]{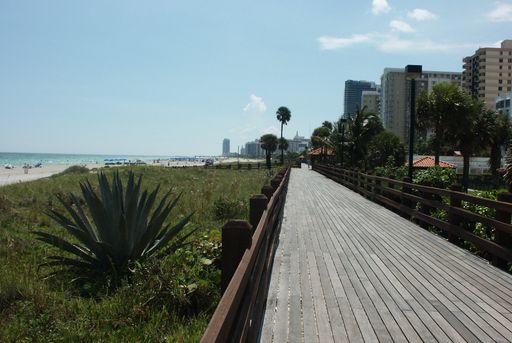}}}
\subfigure[SSIM:0.9585] { {\includegraphics[height=0.9in, width=1.6in]{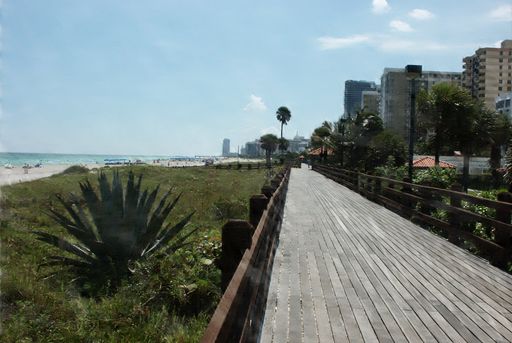}}}
\caption{Comparasion between Baseline and Baseline-RR, (a) rainy image, (c) clean image, (b) is the result of ResGuideNet without Recursion(ResGuideNet-NRecur), (d) is the result of ResGuideNet .}
\end{figure}
\begin{figure*}
 {\includegraphics[height=0.5in, width=0.8in]{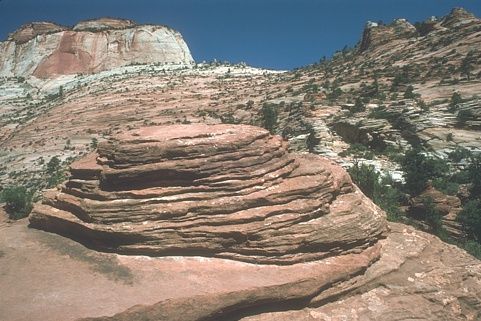}}
 {\includegraphics[height=0.5in, width=0.8in]{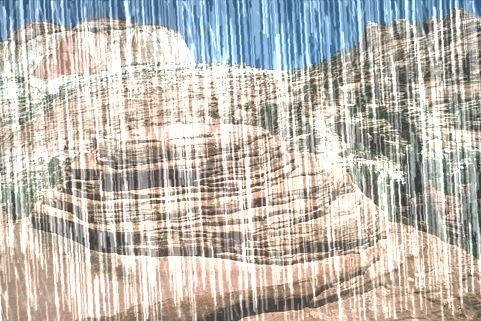}} 
 {\includegraphics[height=0.5in, width=0.8in]{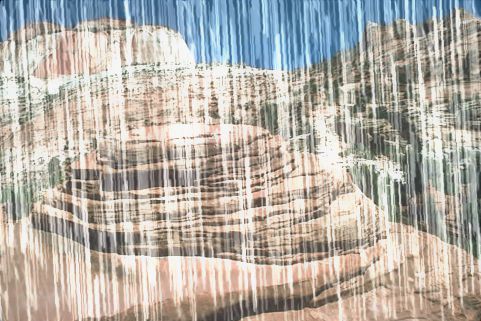}}
 {\includegraphics[height=0.5in, width=0.8in]{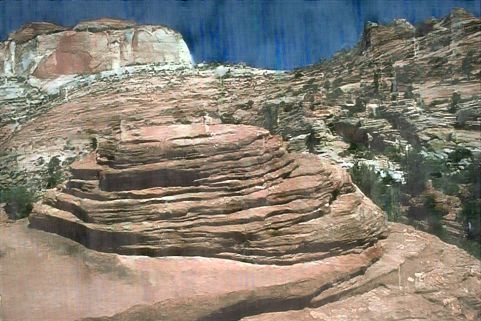}}
 {\includegraphics[height=0.5in, width=0.8in]{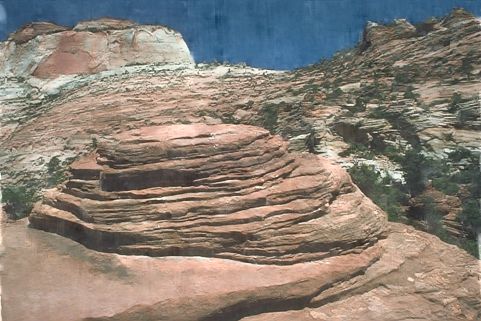}} 
 {\includegraphics[height=0.5in, width=0.8in]{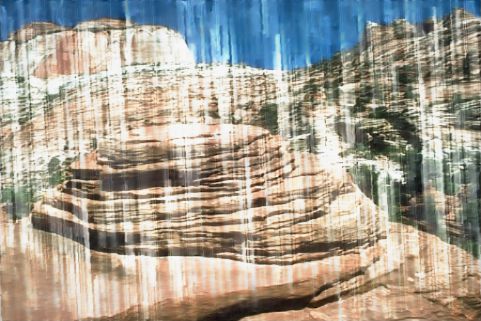}} 
 {\includegraphics[height=0.5in, width=0.8in]{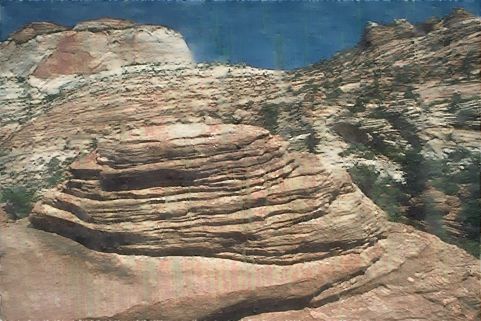}}
 {\includegraphics[height=0.5in, width=0.8in]{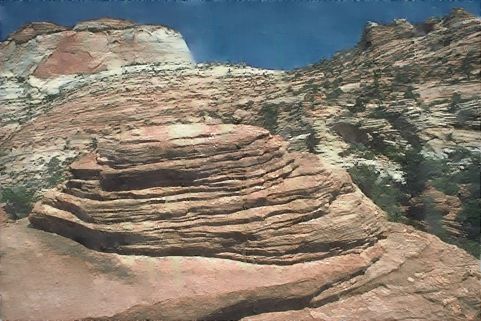}}
 {\includegraphics[height=0.5in, width=0.8in]{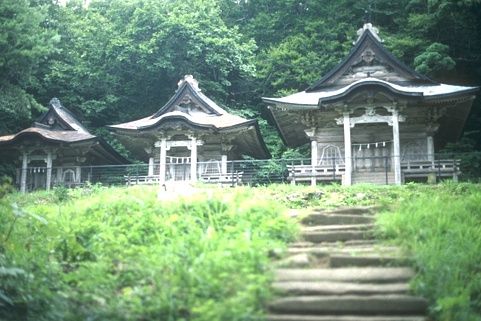}}
 {\includegraphics[height=0.5in, width=0.8in]{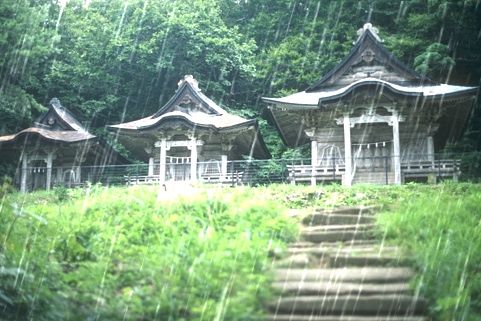}} 
 {\includegraphics[height=0.5in, width=0.8in]{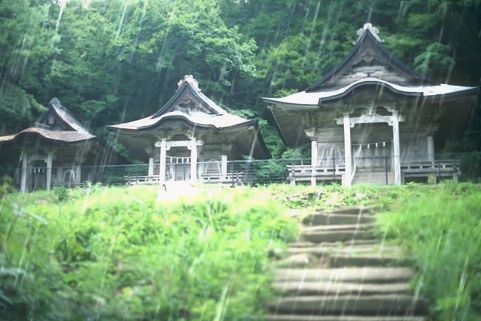}}
 {\includegraphics[height=0.5in, width=0.8in]{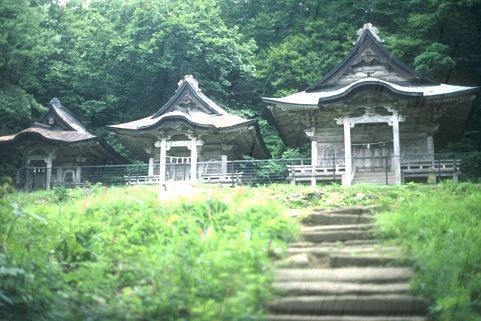}}
 {\includegraphics[height=0.5in, width=0.8in]{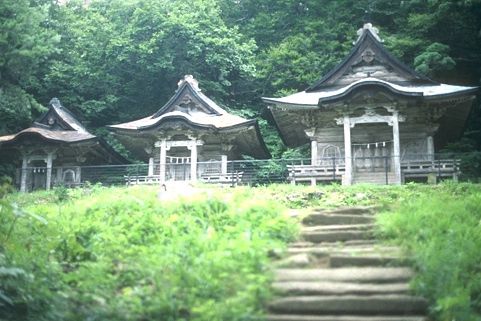}} 
  {\includegraphics[height=0.5in, width=0.8in]{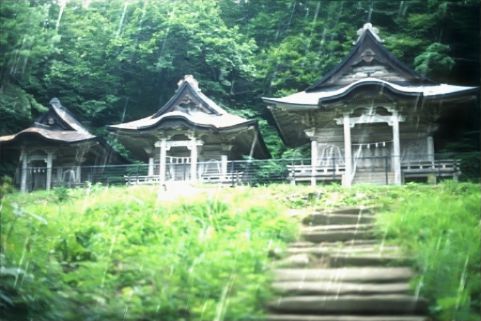}} 
 {\includegraphics[height=0.5in, width=0.8in]{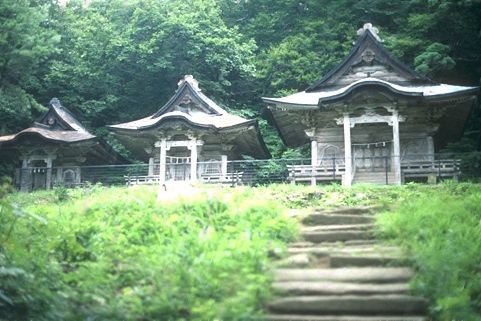}}
 {\includegraphics[height=0.5in, width=0.8in]{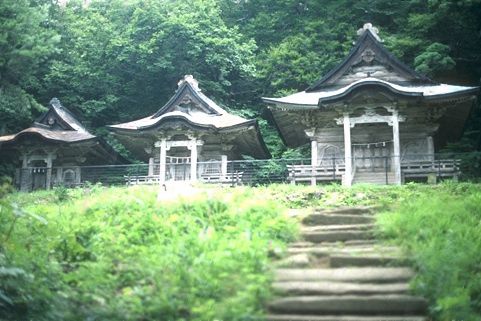}} \\ \vspace{-0.07in}
\subfigure[Clean image] {\includegraphics[height=0.5in, width=0.8in]{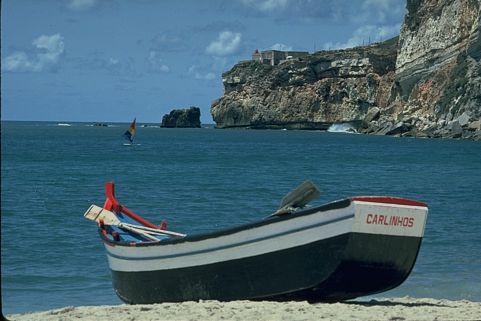}}
\subfigure[Rainy imag] {\includegraphics[height=0.5in, width=0.8in]{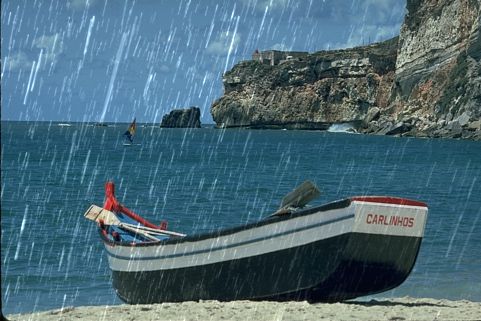}} 
\subfigure[GMM] {\includegraphics[height=0.5in, width=0.8in]{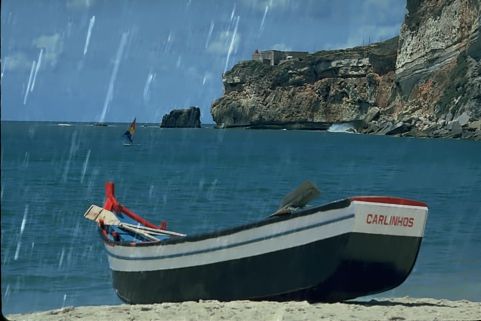}}
\subfigure[DDN] {\includegraphics[height=0.5in, width=0.8in]{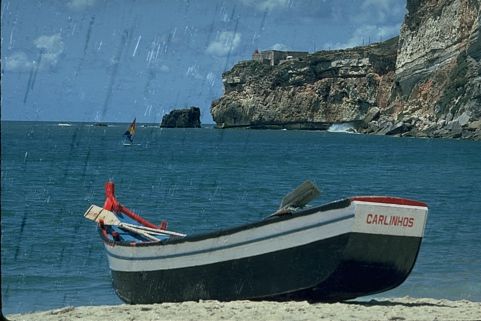}}
\subfigure[JORDER]  {\includegraphics[height=0.5in, width=0.8in]{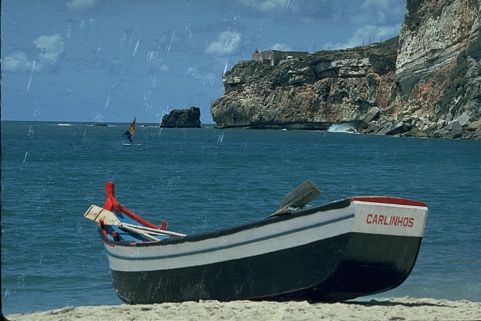}} 
\subfigure[DID-MDN] {\includegraphics[height=0.5in, width=0.8in]{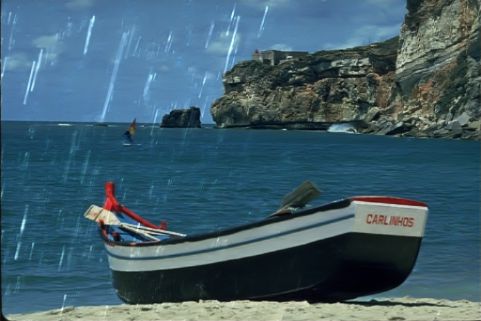}} 
\subfigure[Ours B$_3$] {\includegraphics[height=0.5in, width=0.8in]{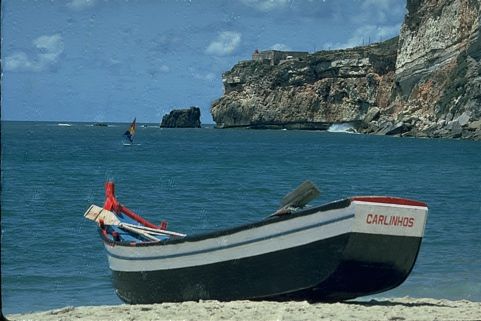}}
\subfigure[ResGuideNet] {\includegraphics[height=0.5in, width=0.8in]{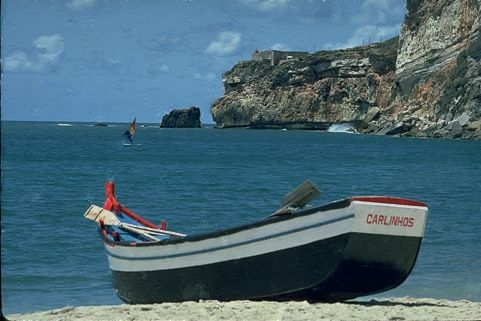}}

\caption{Three results on synthetic images.}
\end{figure*}

\begin{figure*}

 {\includegraphics[height=0.7in, width=1.1in]{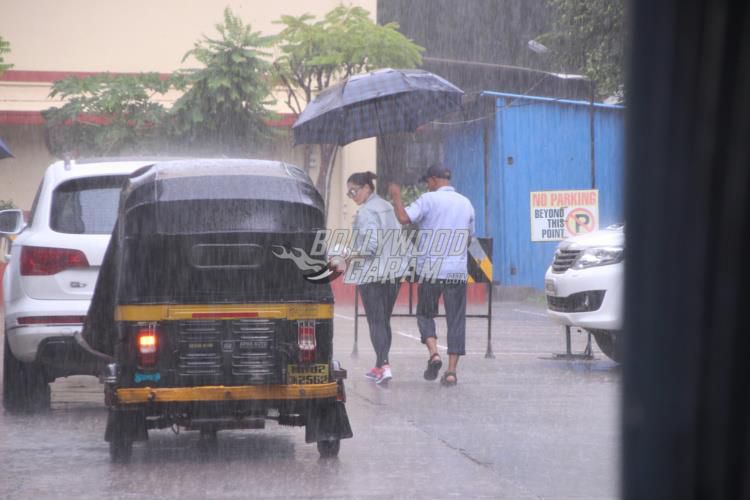}}
 {\includegraphics[height=0.7in, width=1.1in]{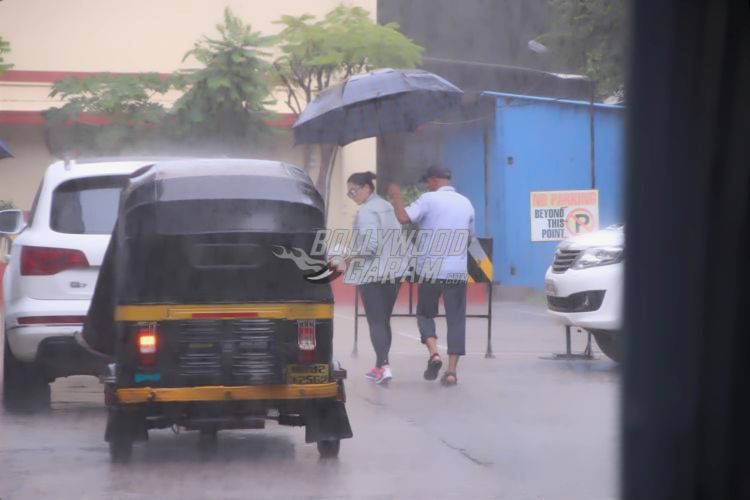}} 
 {\includegraphics[height=0.7in, width=1.1in]{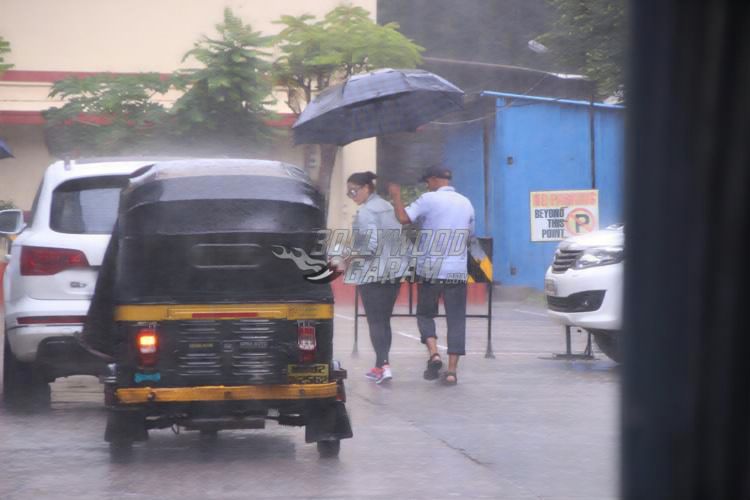}}
 {\includegraphics[height=0.7in, width=1.1in]{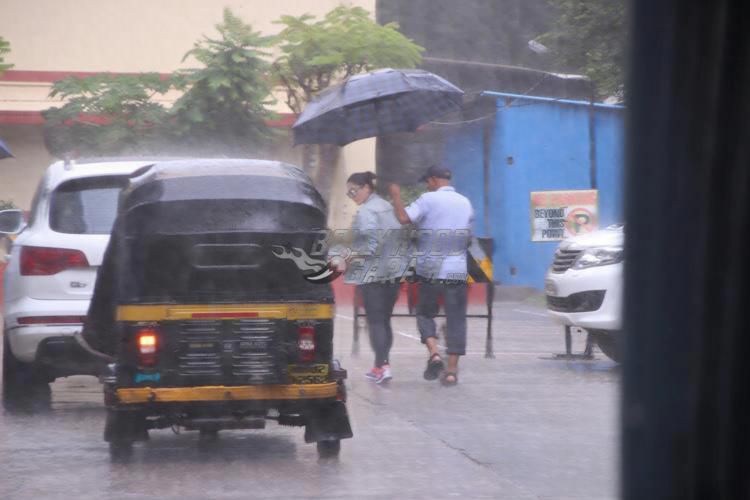}}
 {\includegraphics[height=0.7in, width=1.1in]{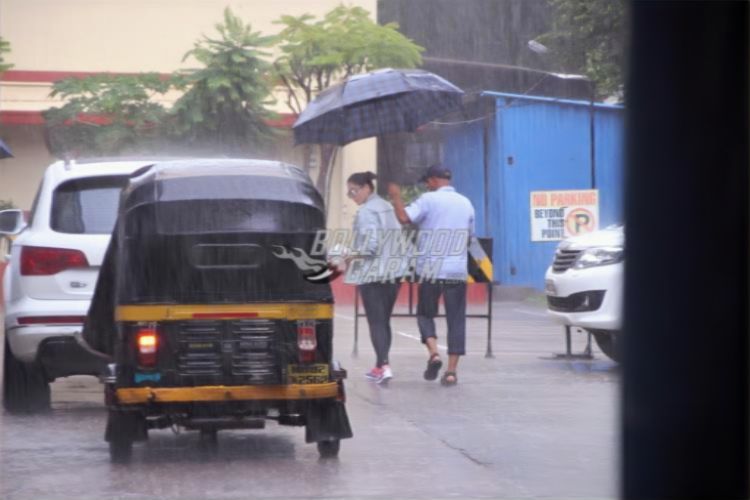}} 
 {\includegraphics[height=0.7in, width=1.1in]{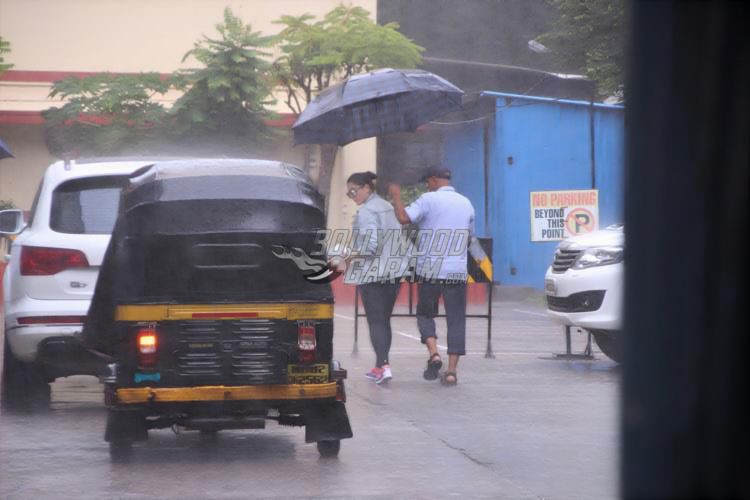}}

 {\includegraphics[height=0.7in, width=1.1in]{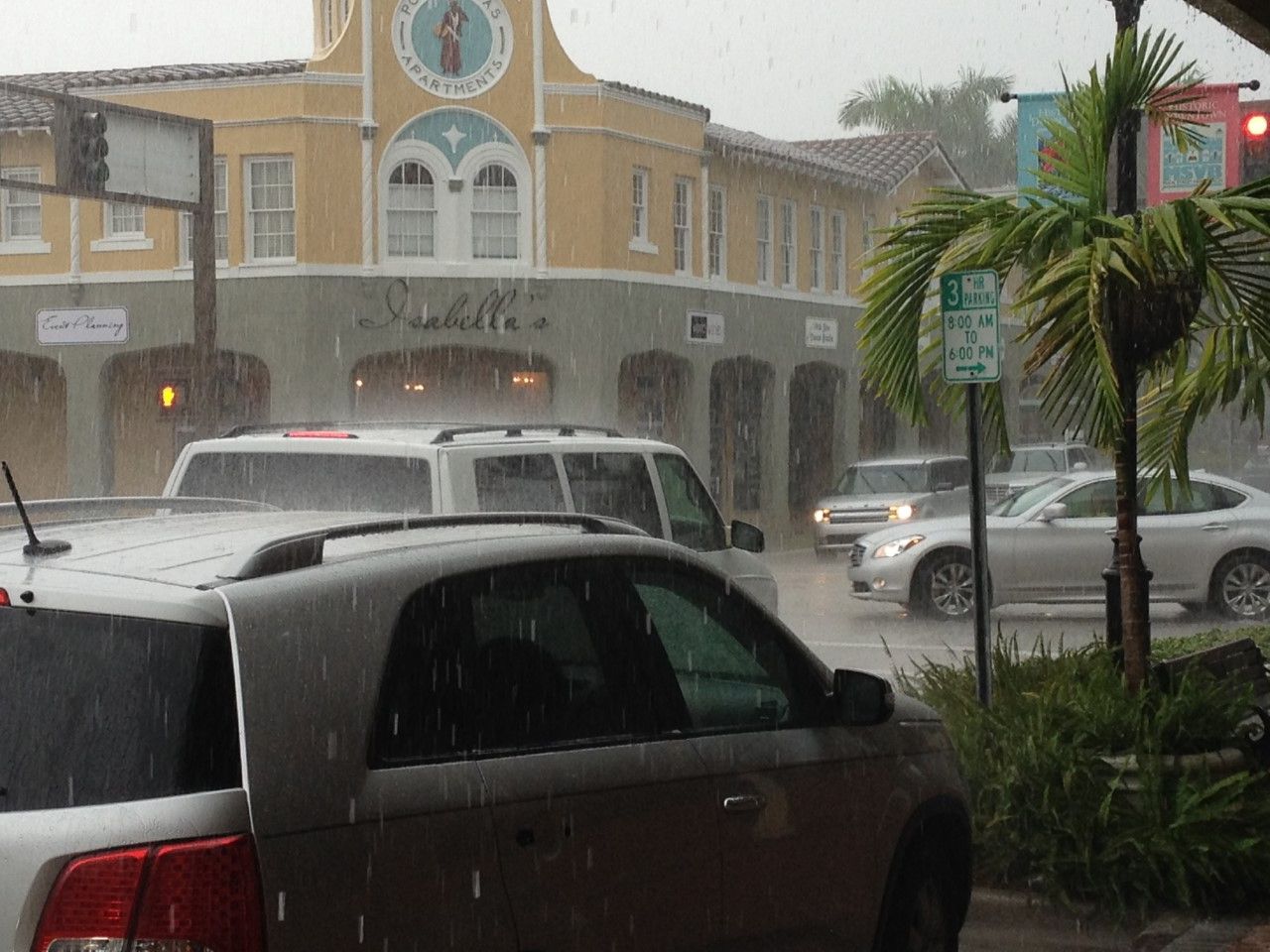}}
 {\includegraphics[height=0.7in, width=1.1in]{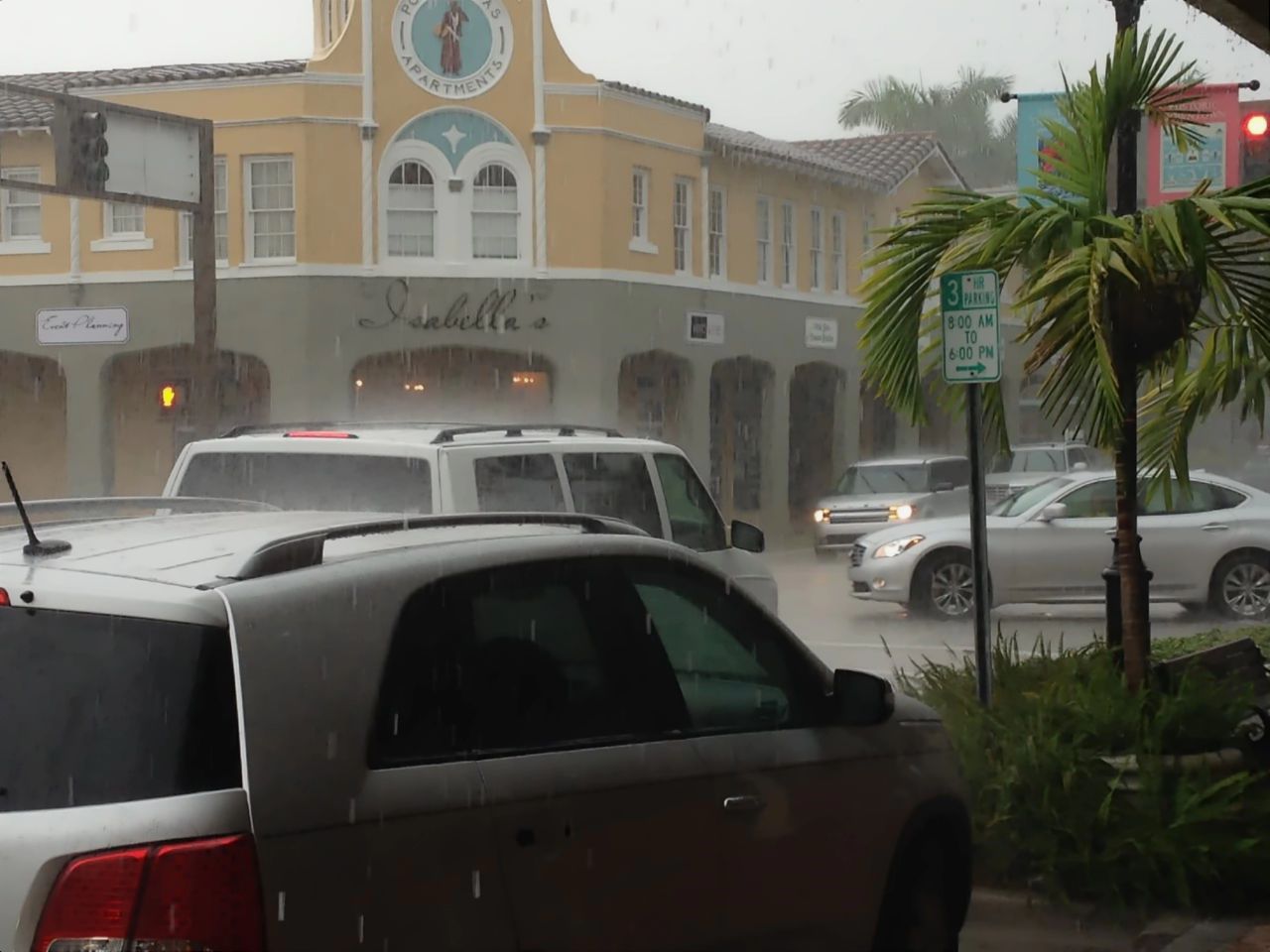}} 
 {\includegraphics[height=0.7in, width=1.1in]{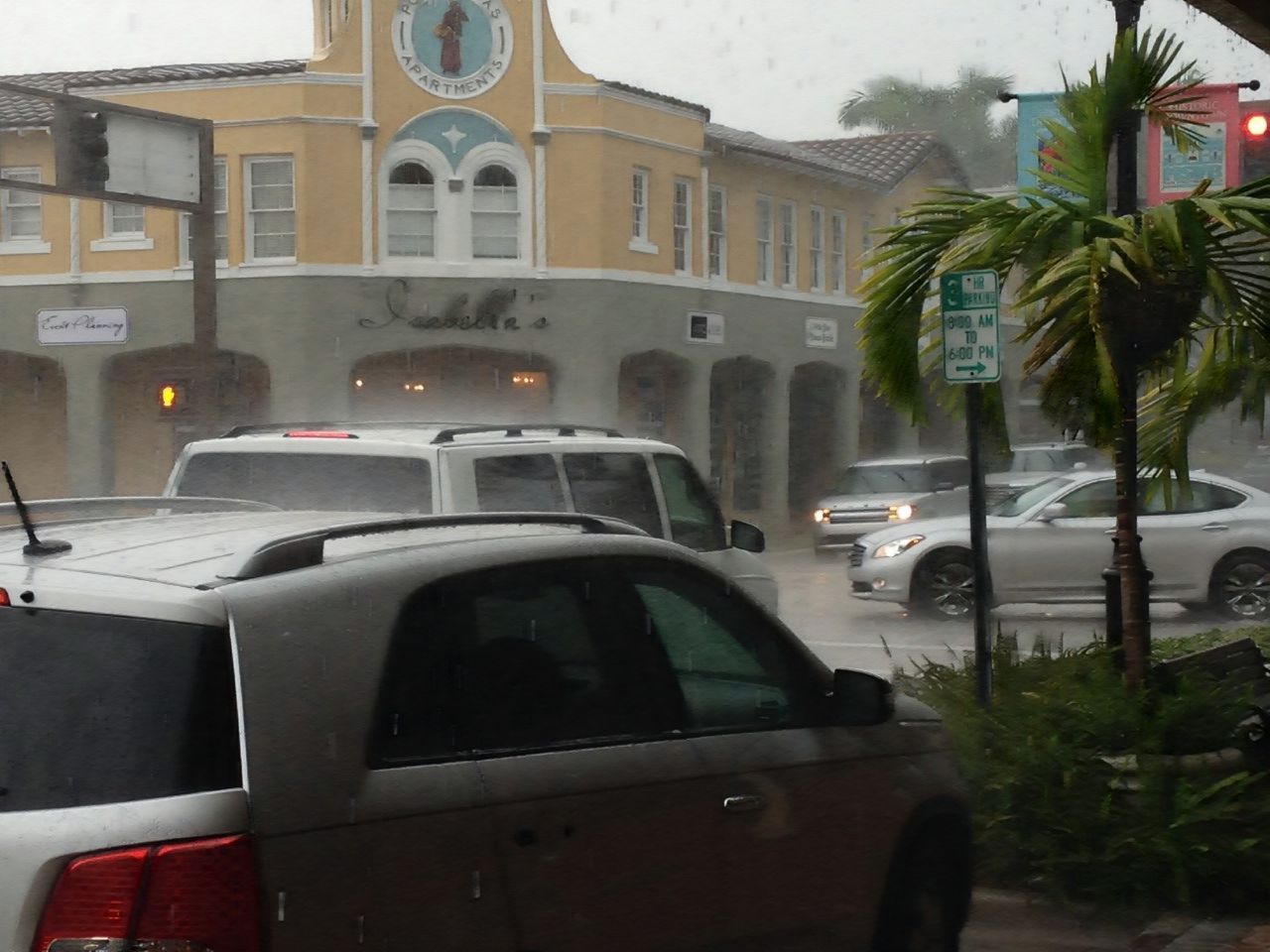}}
 {\includegraphics[height=0.7in, width=1.1in]{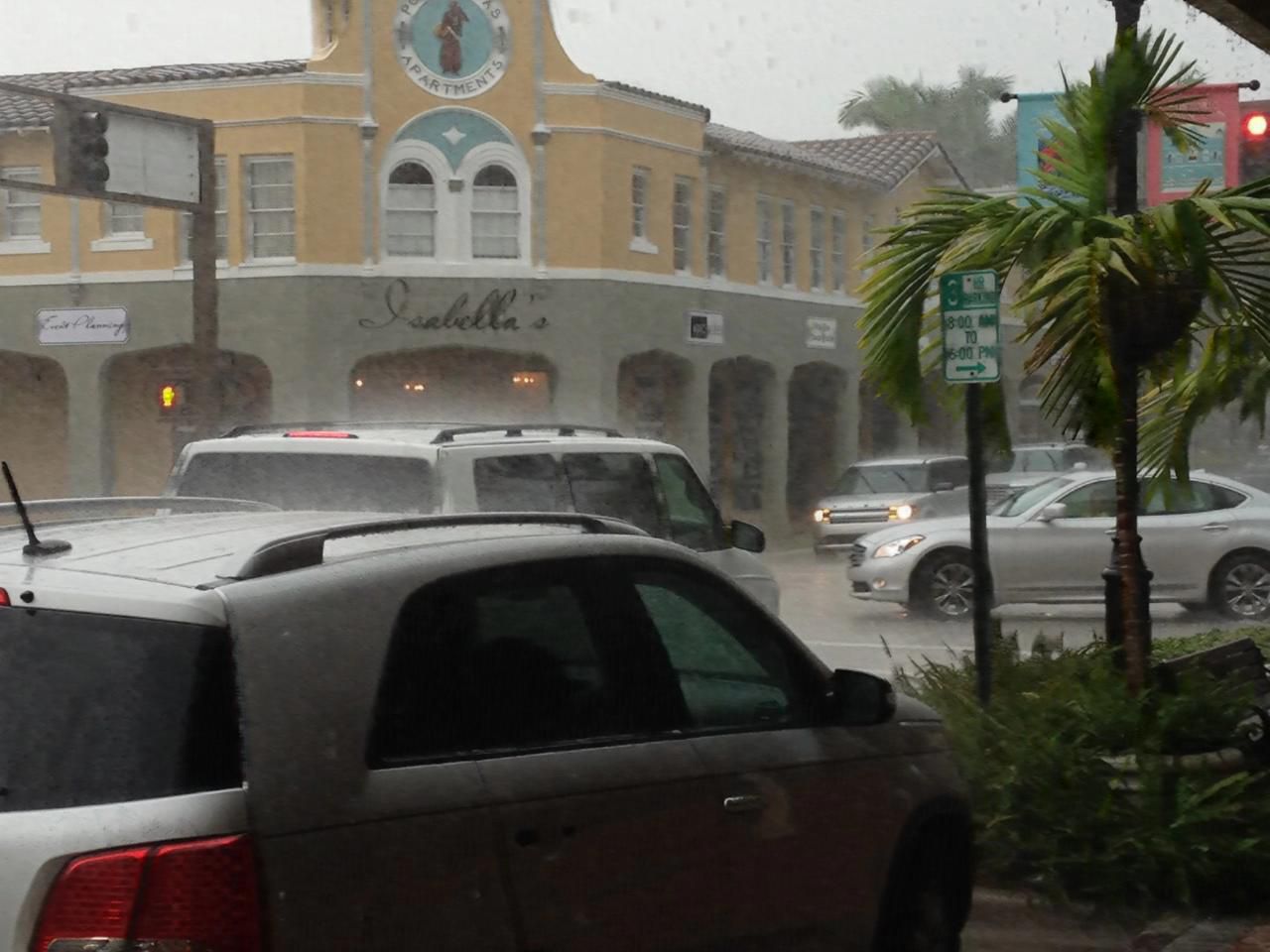}}
 {\includegraphics[height=0.7in, width=1.1in]{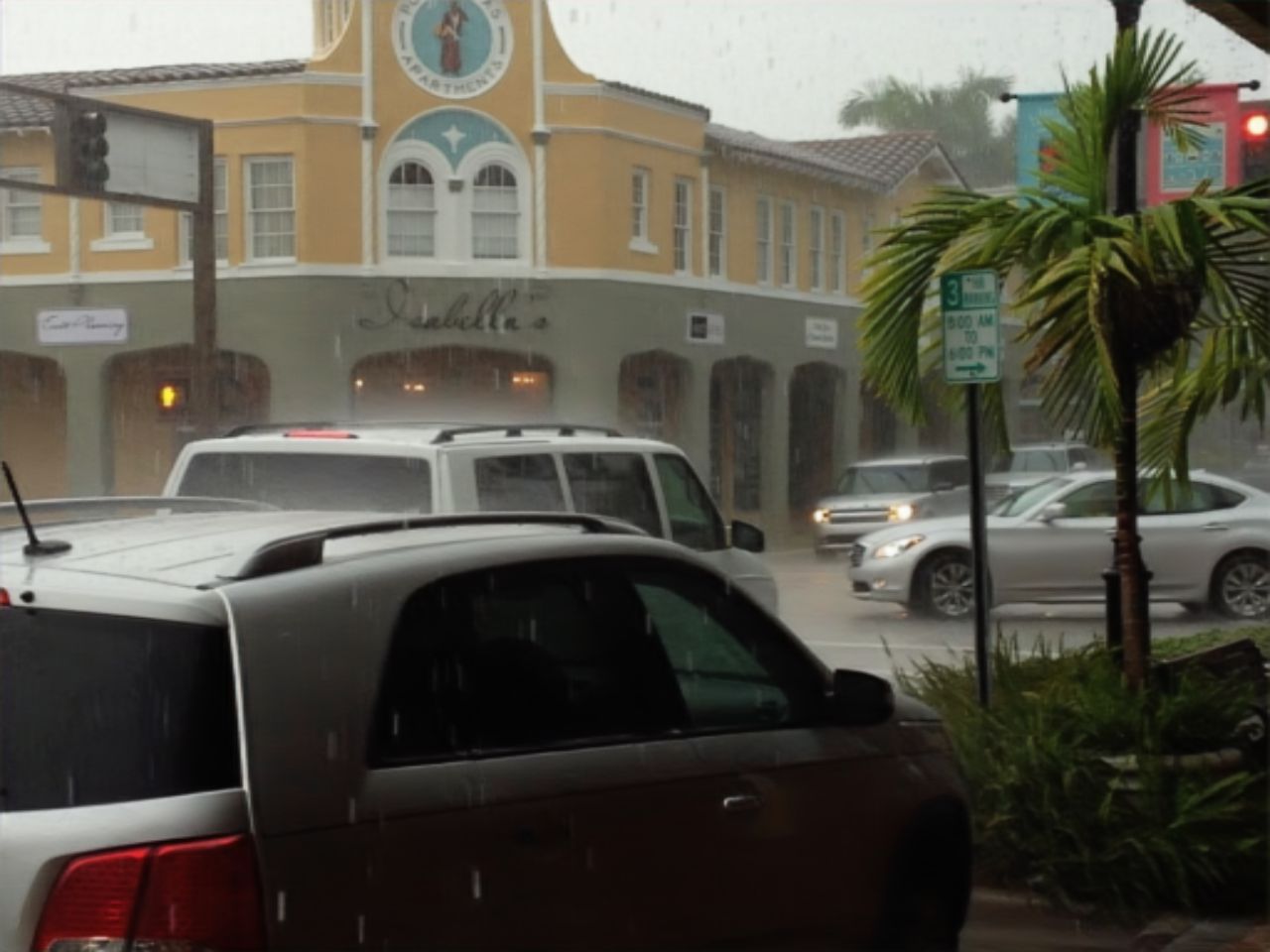}} 
 {\includegraphics[height=0.7in, width=1.1in]{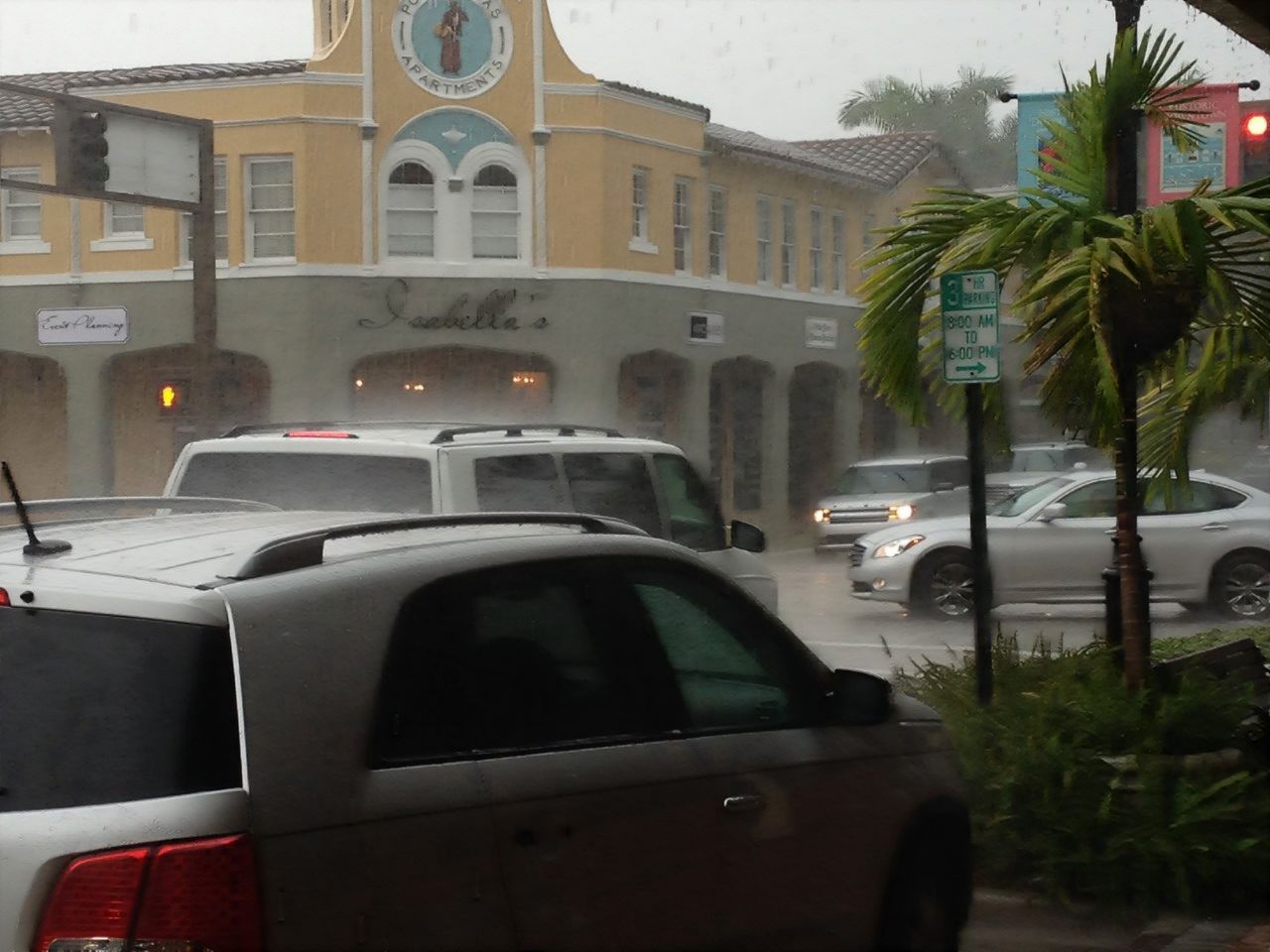}}

 {\includegraphics[height=0.7in, width=1.1in]{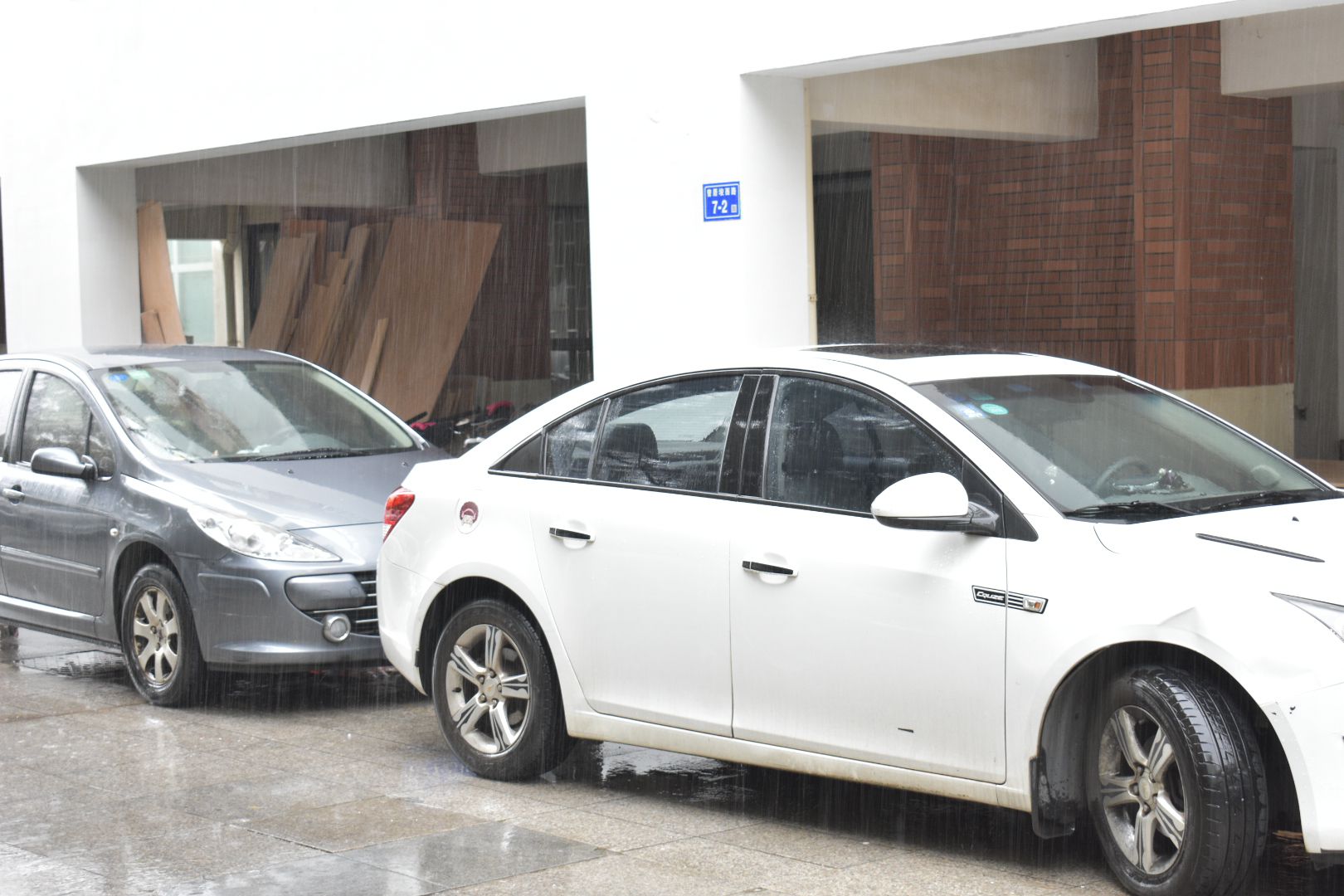}}
 {\includegraphics[height=0.7in, width=1.1in]{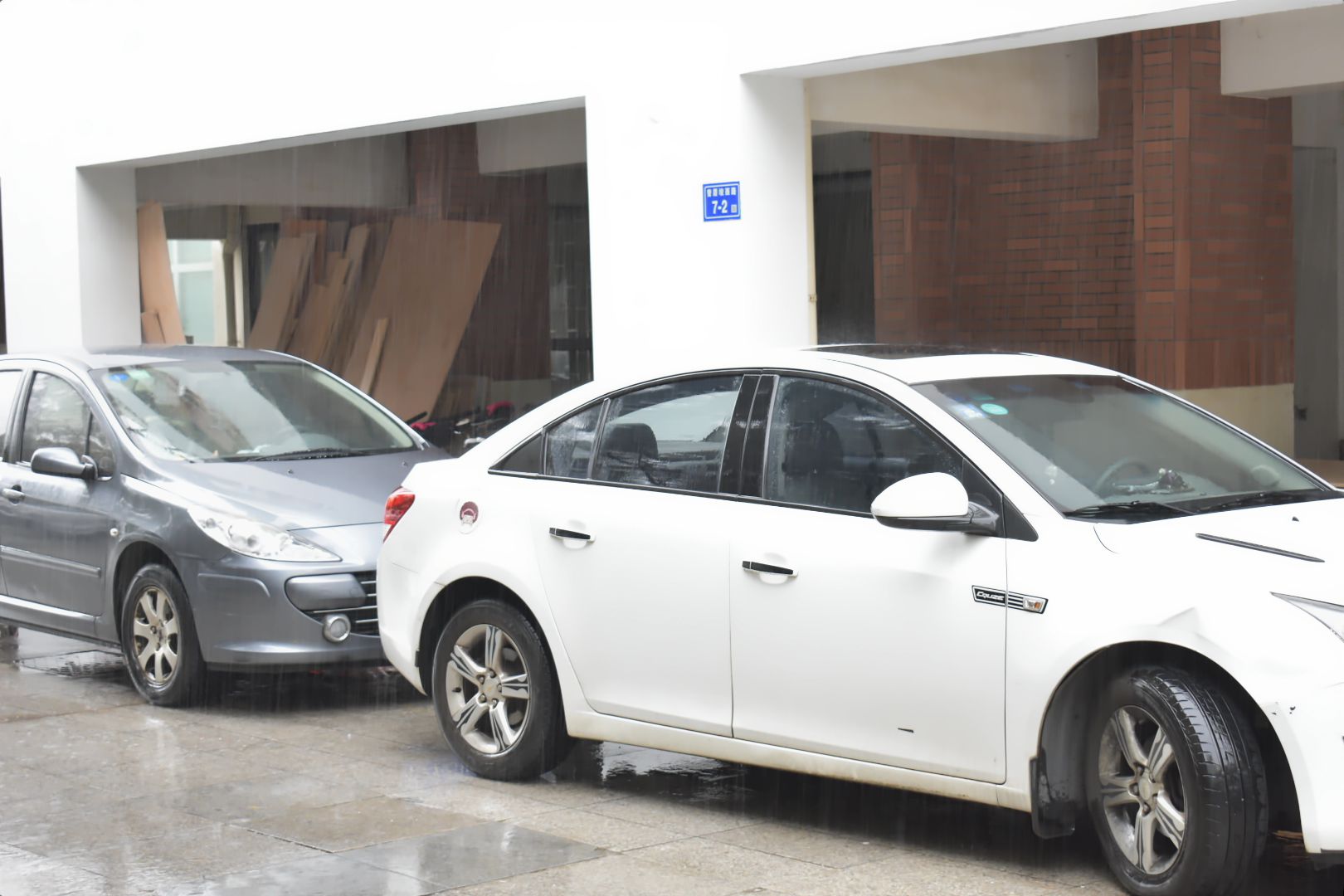}} 
 {\includegraphics[height=0.7in, width=1.1in]{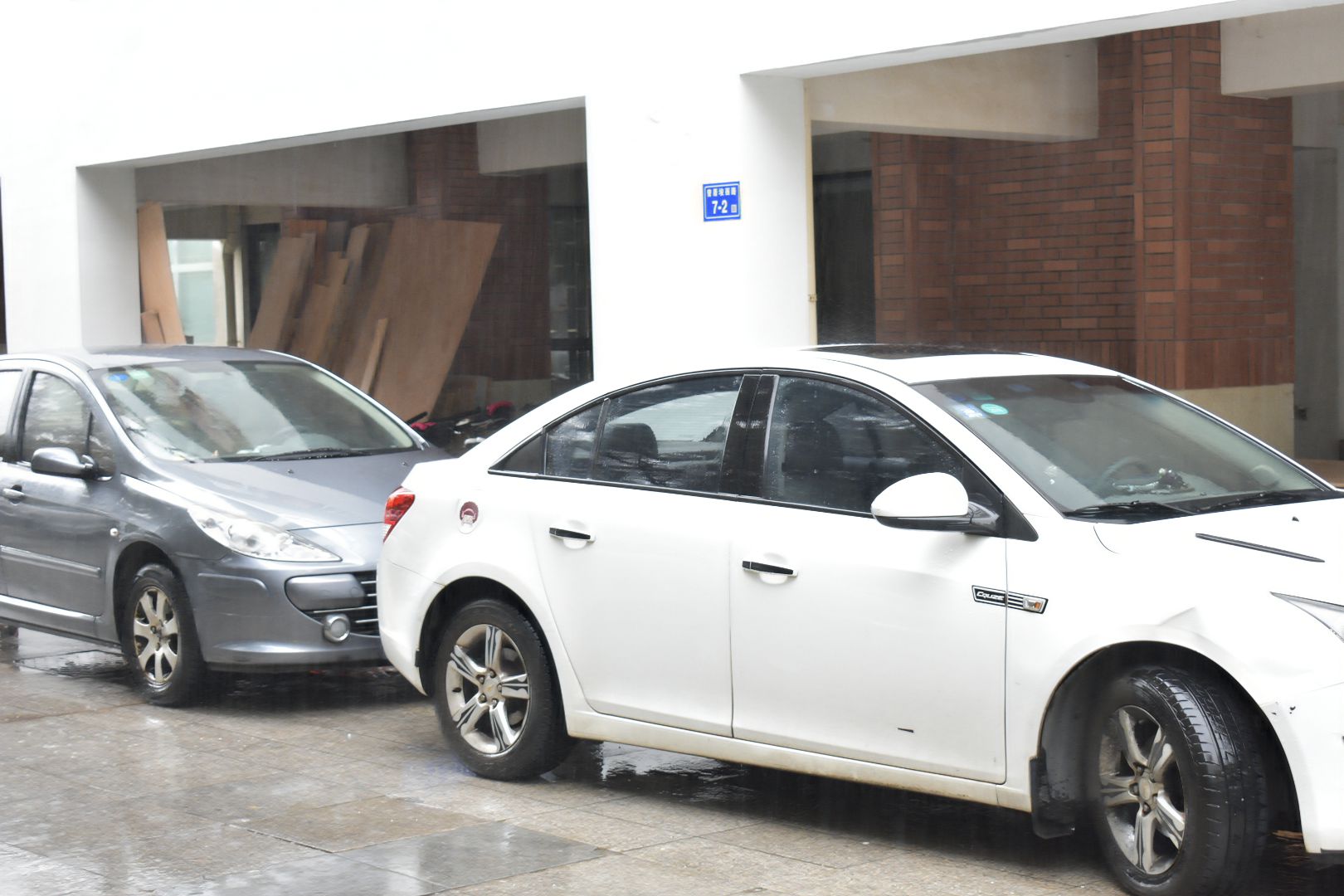}}
 {\includegraphics[height=0.7in, width=1.1in]{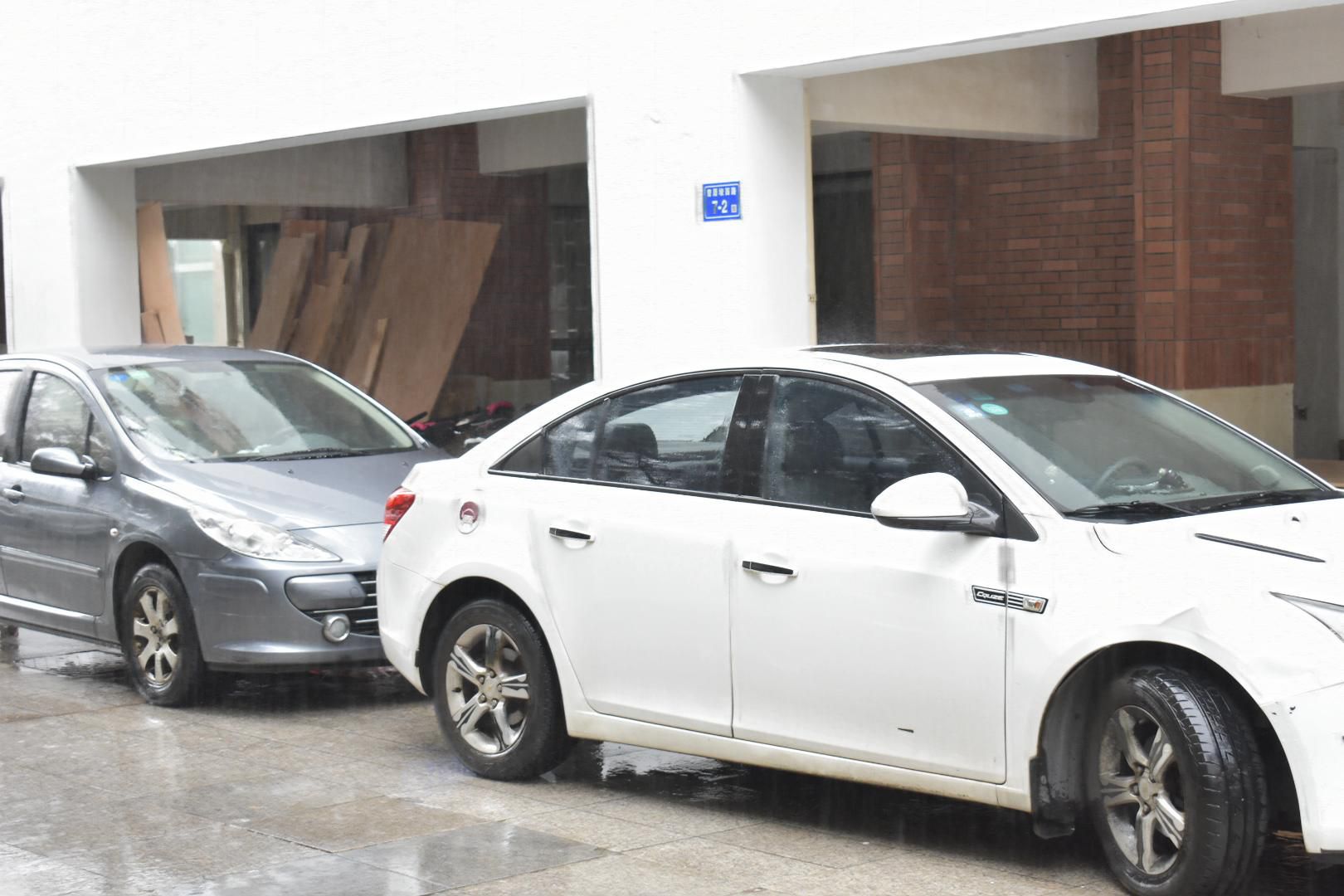}}
 {\includegraphics[height=0.7in, width=1.1in]{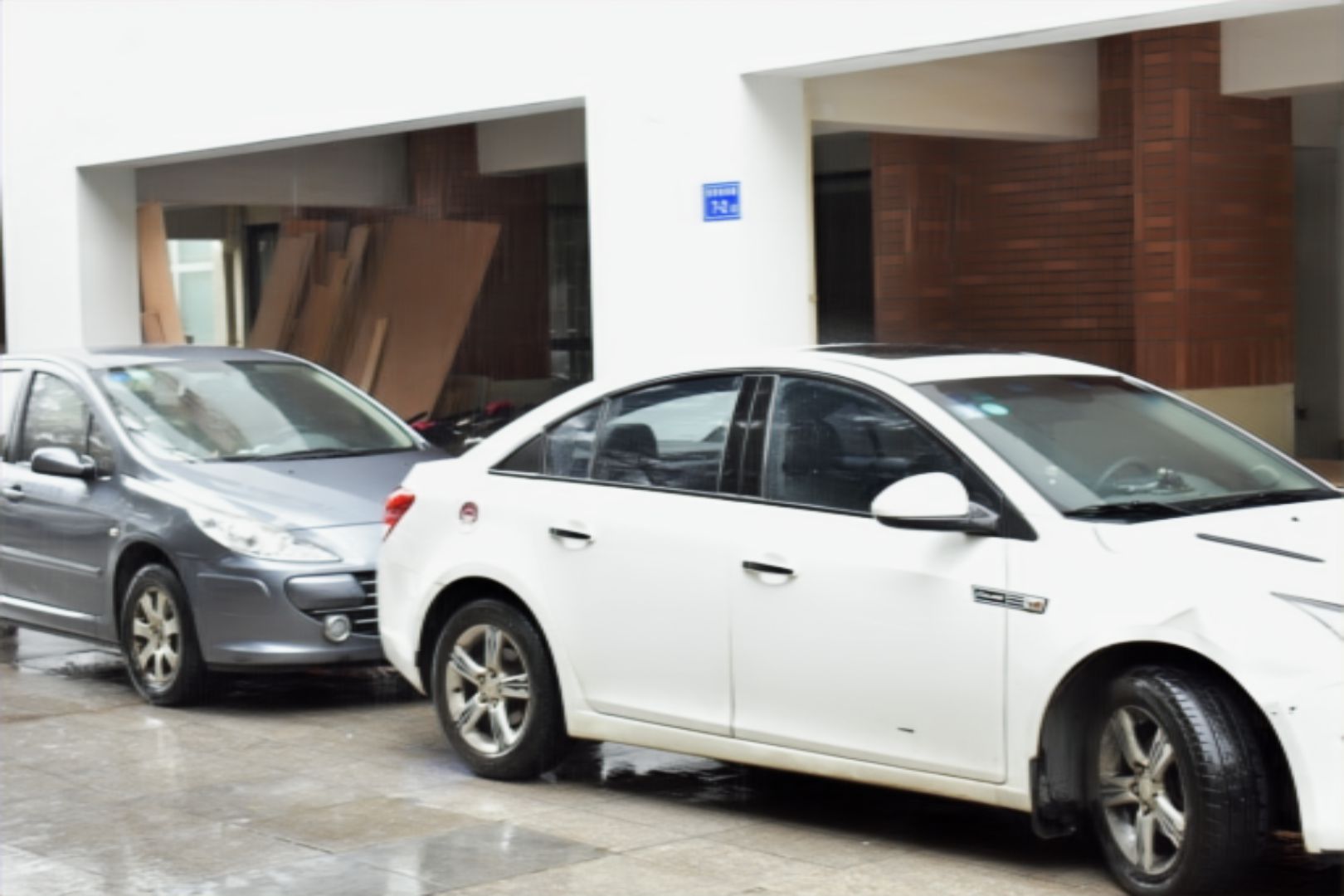}} 
 {\includegraphics[height=0.7in, width=1.1in]{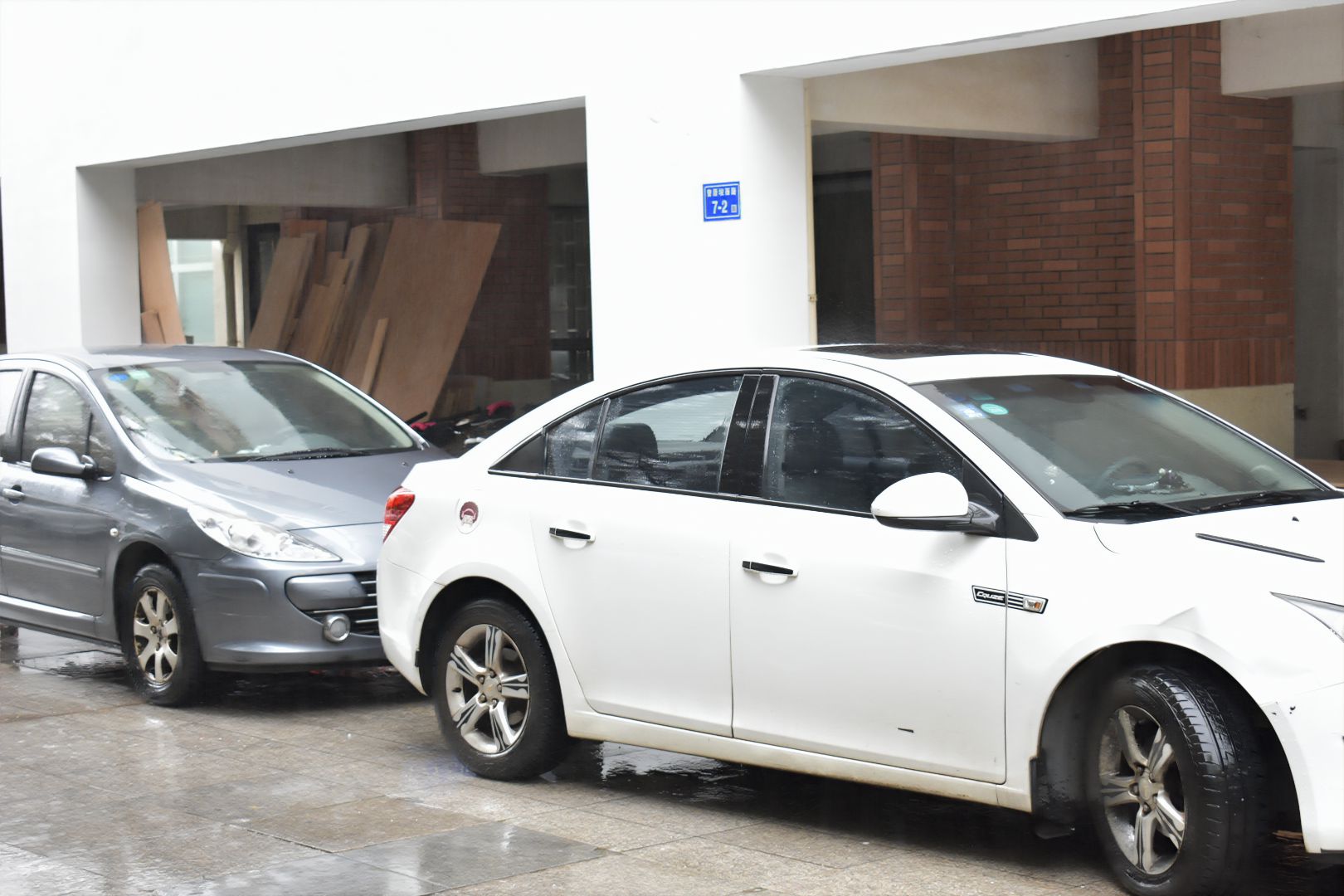}}

{\includegraphics[height=0.7in, width=1.1in]{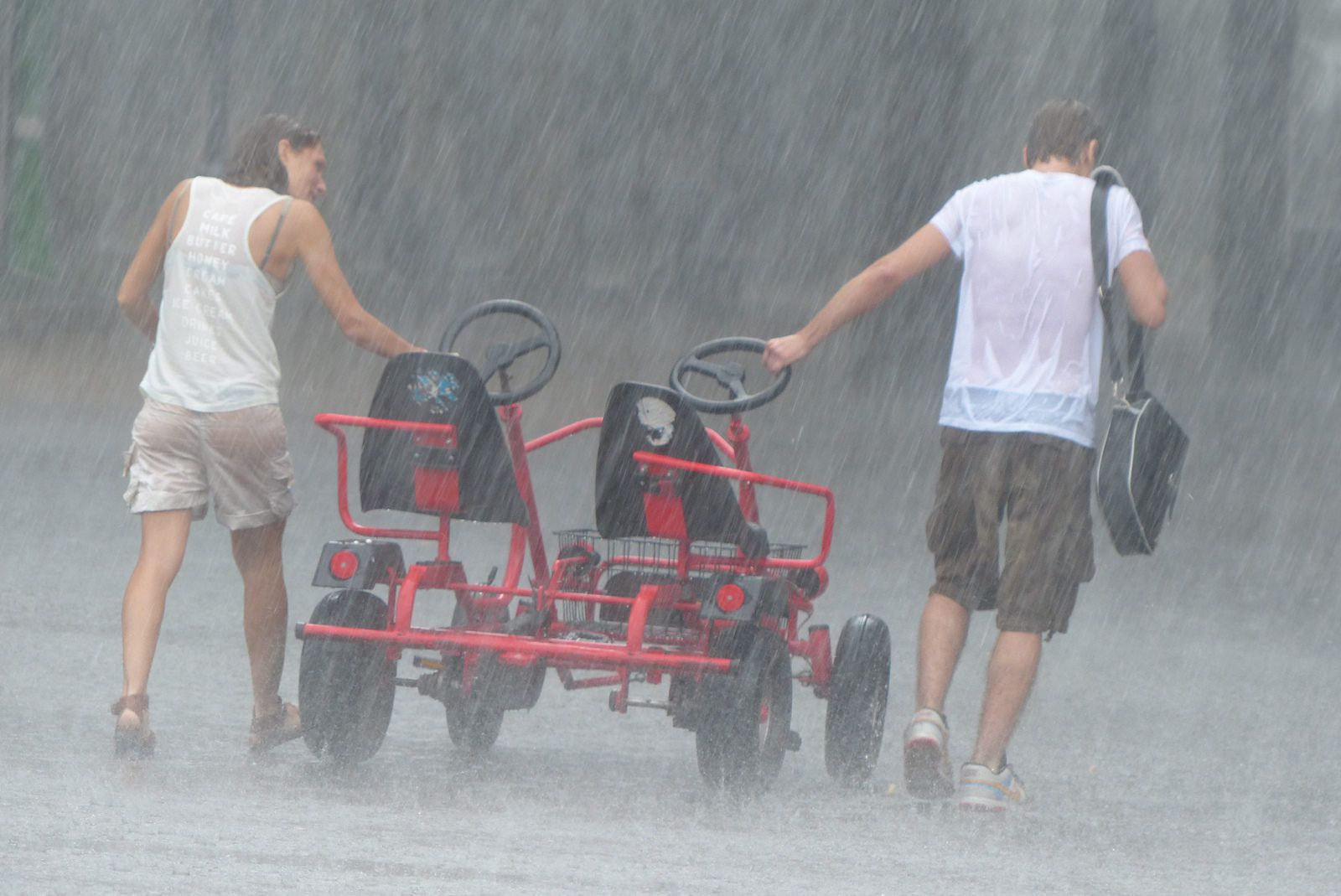}}
 {\includegraphics[height=0.7in, width=1.1in]{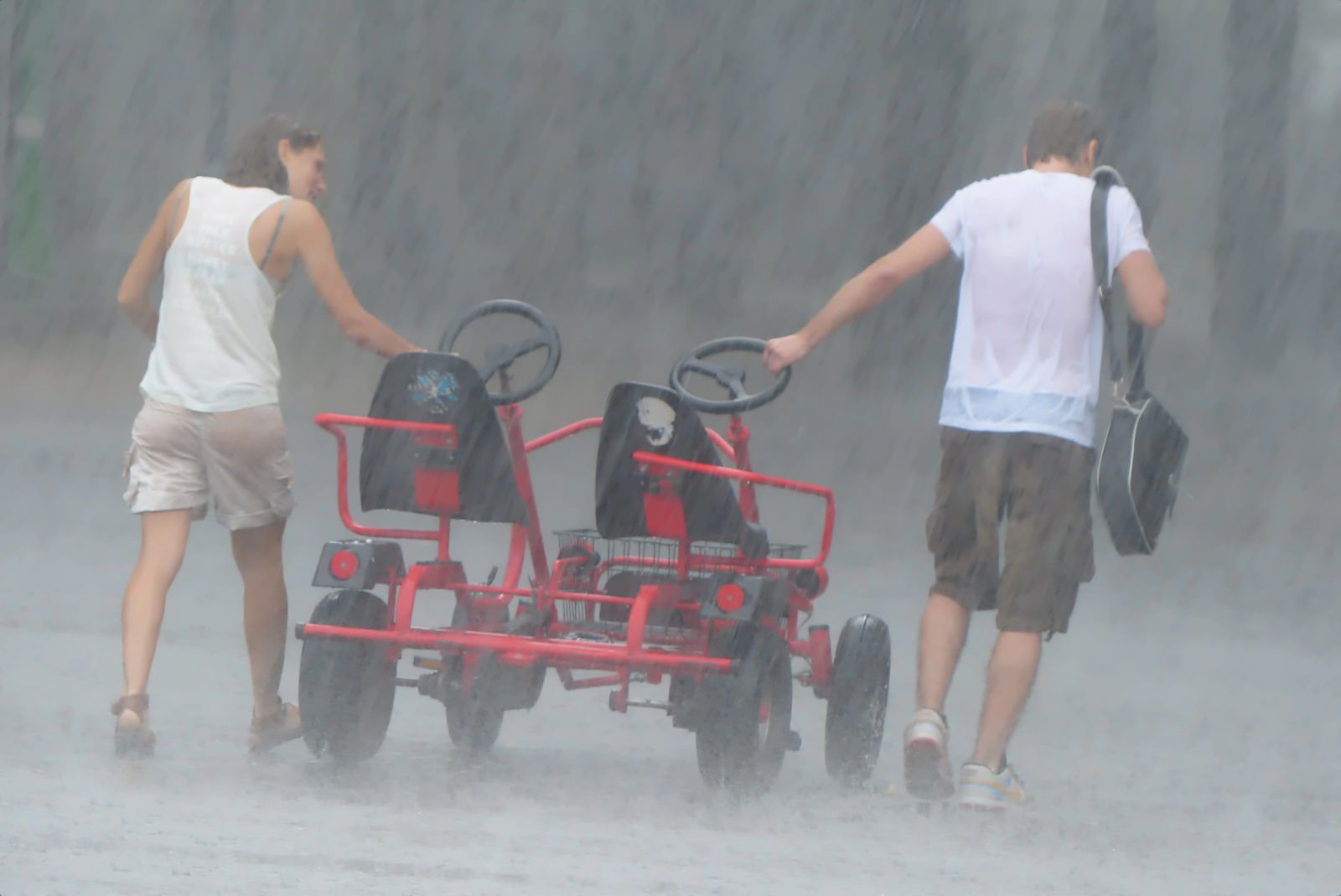}} 
  {\includegraphics[height=0.7in, width=1.1in]{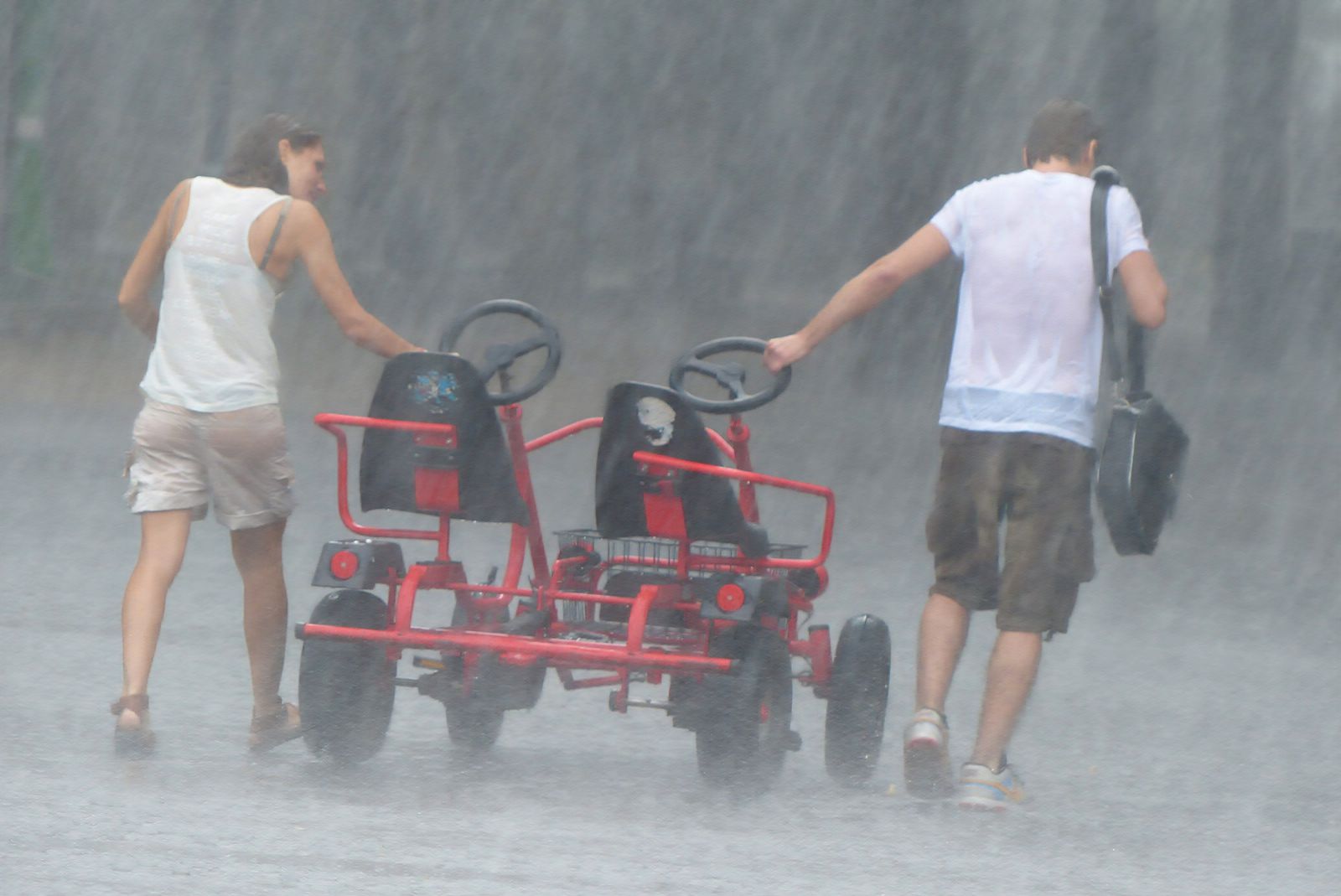}}
 {\includegraphics[height=0.7in, width=1.1in]{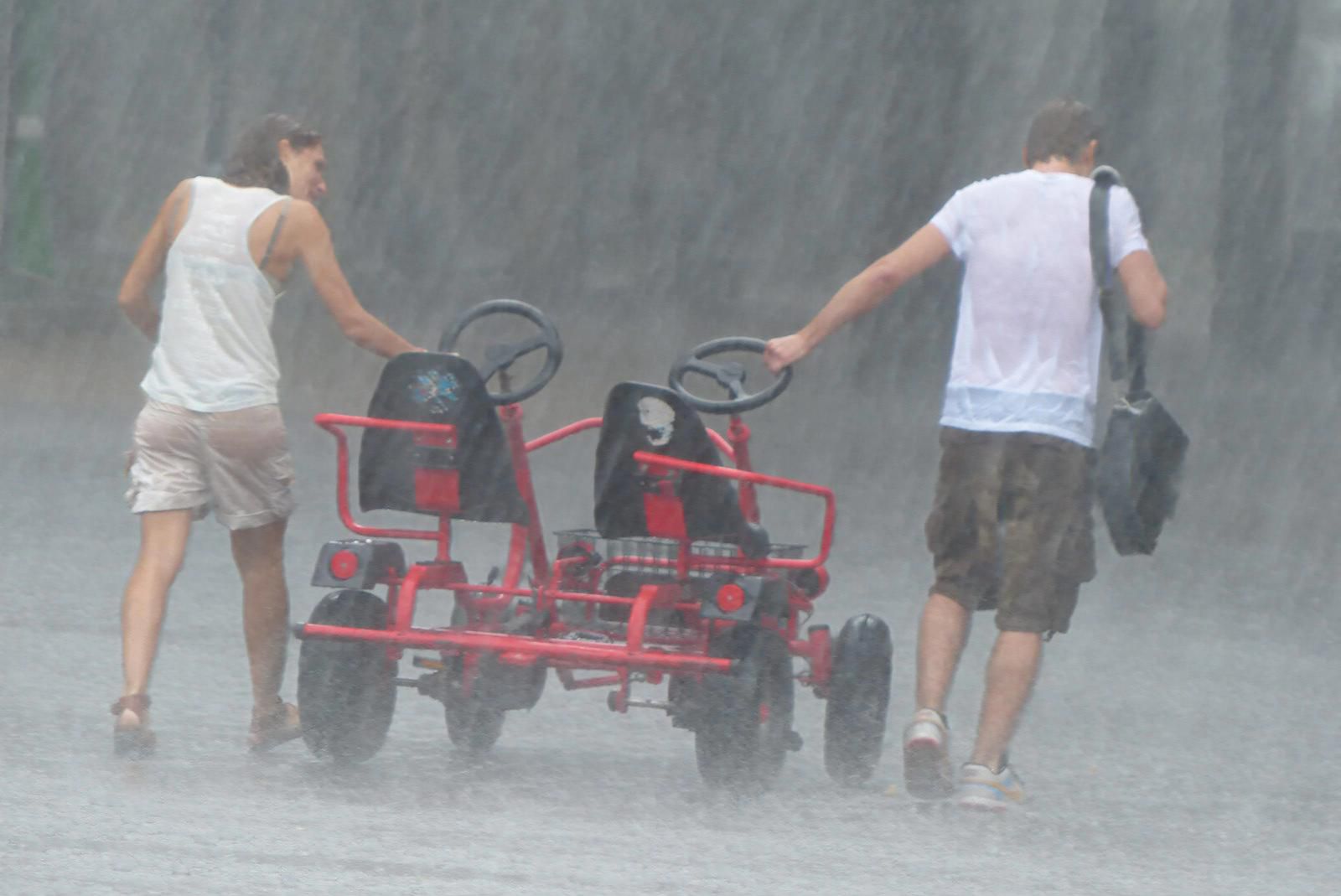}}
{\includegraphics[height=0.7in, width=1.1in]{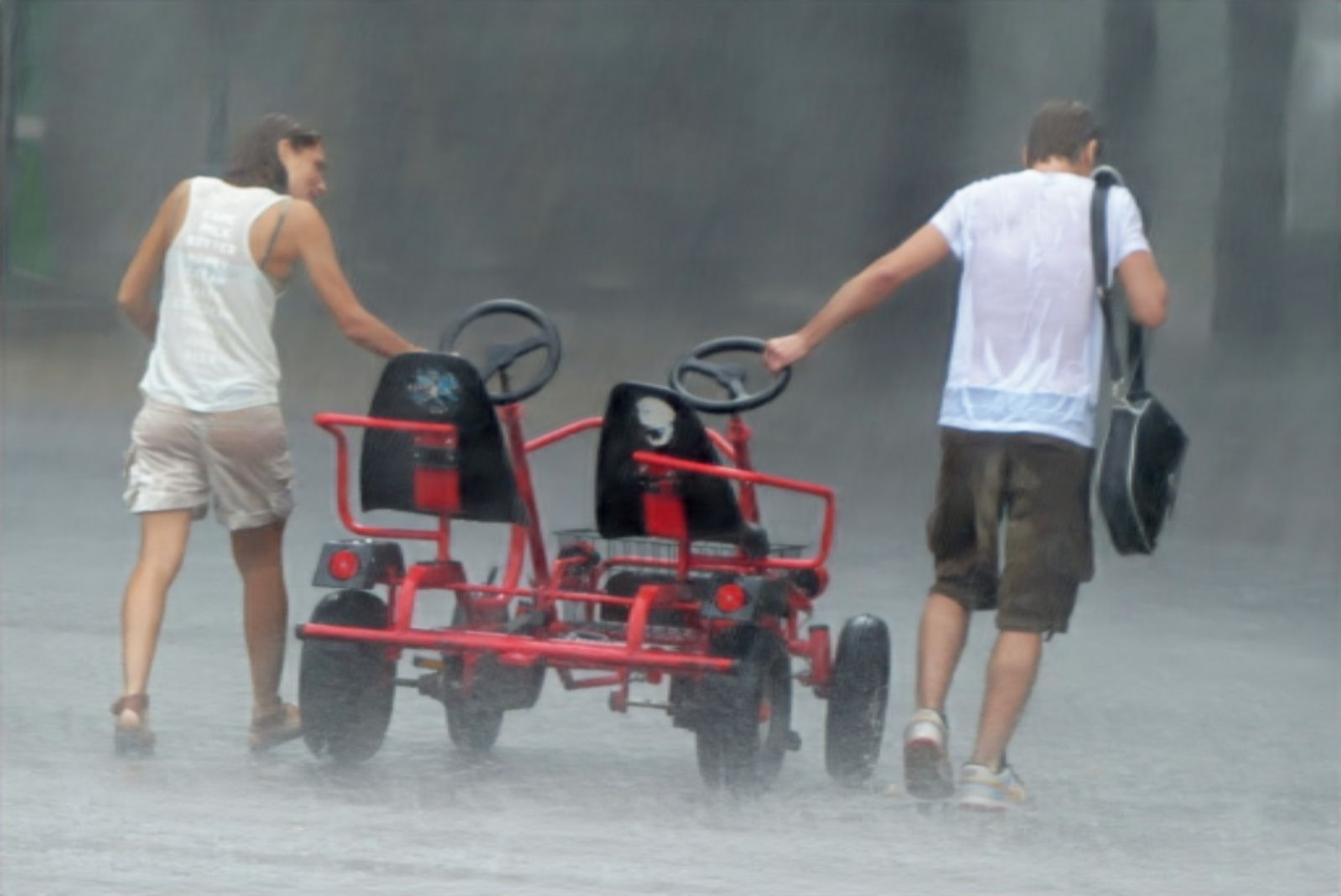}} 
  {\includegraphics[height=0.7in, width=1.1in]{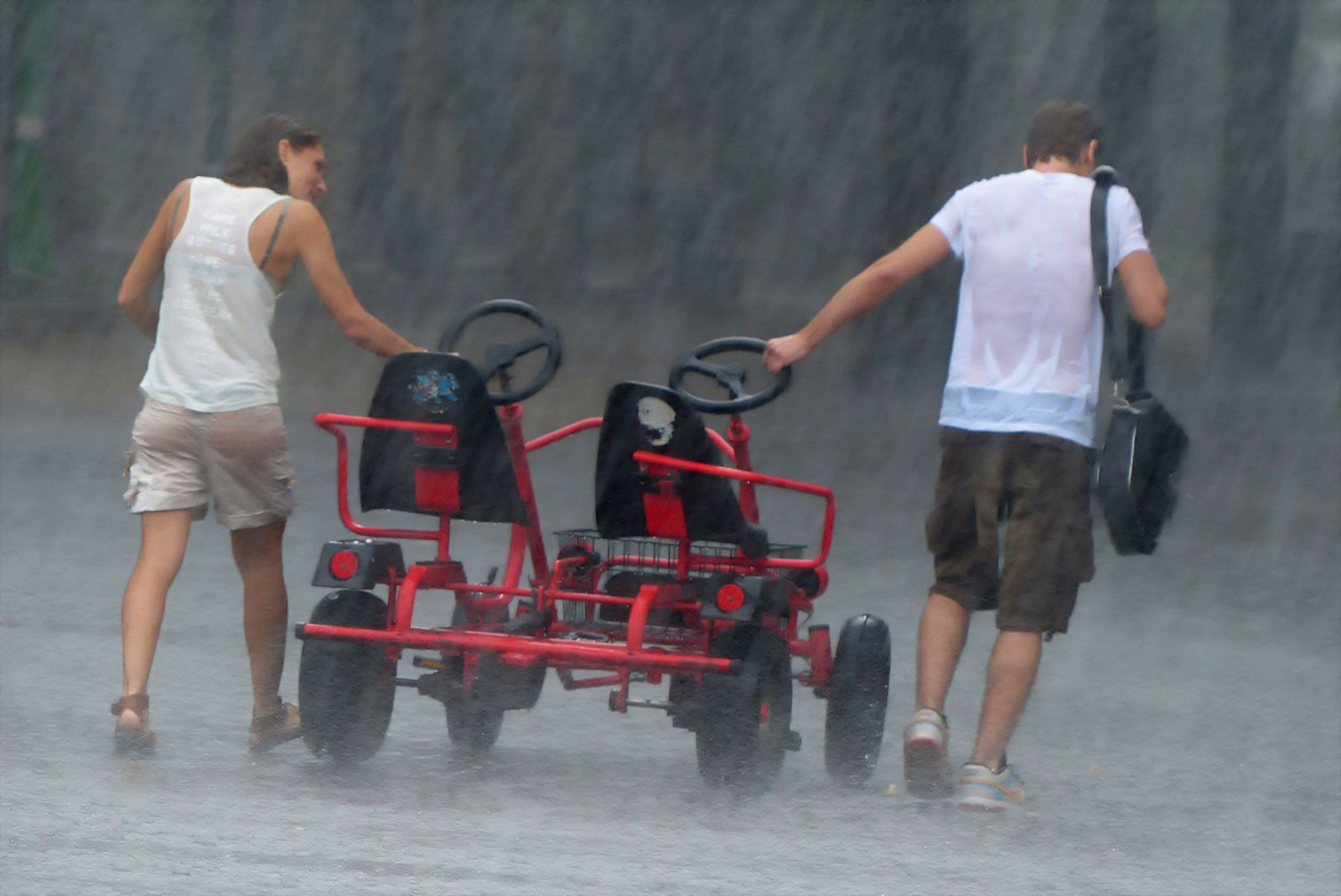}} \\ \vspace{-0.07in}

 \subfigure[Rainy image]{\includegraphics[height=0.7in, width=1.1in]{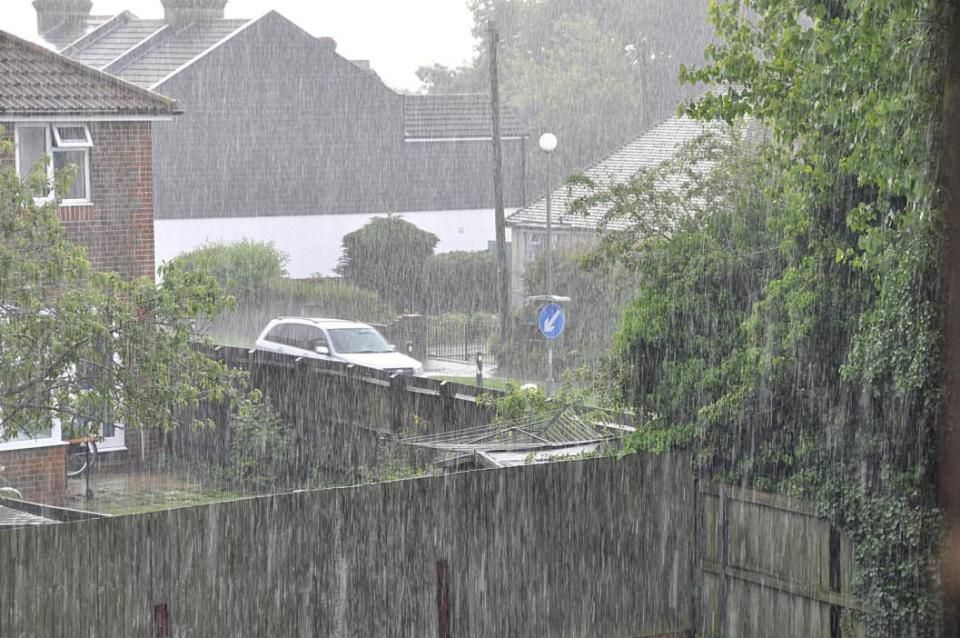}}
 \subfigure[GMM]{\includegraphics[height=0.7in, width=1.1in]{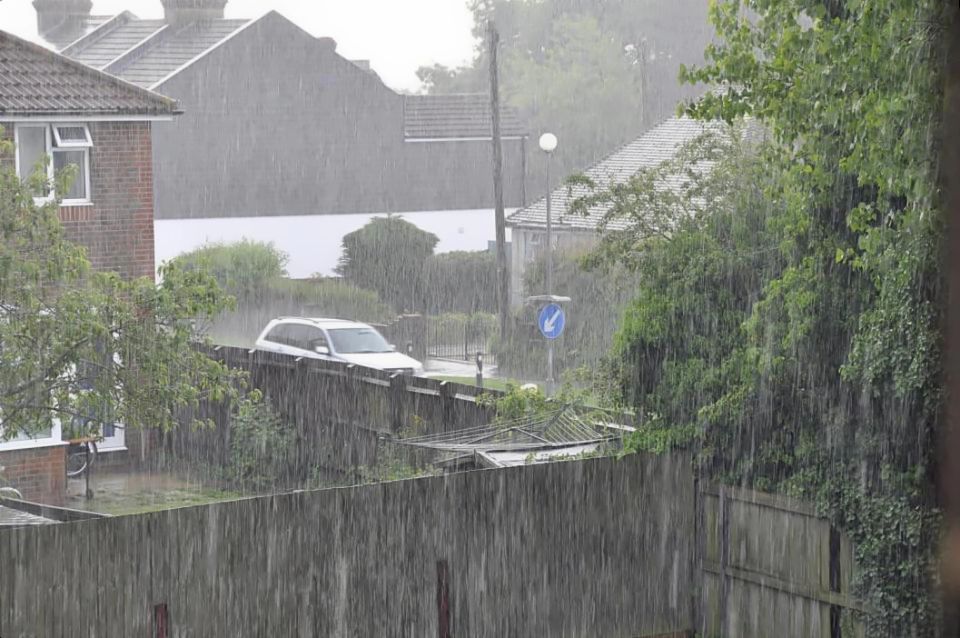}} 
\subfigure[DDN] {\includegraphics[height=0.7in, width=1.1in]{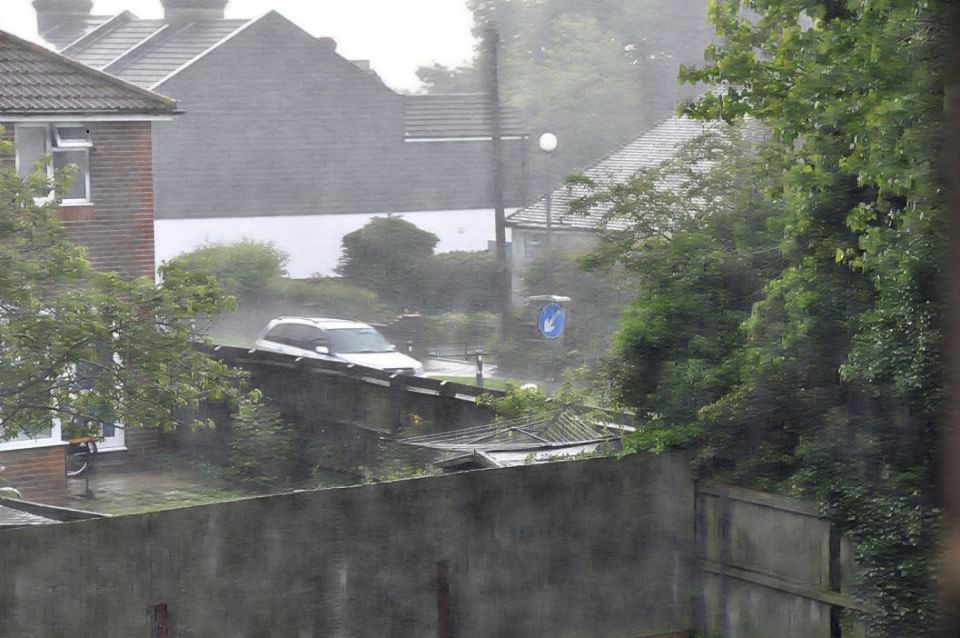}}
 \subfigure[JORDER]{\includegraphics[height=0.7in, width=1.1in]{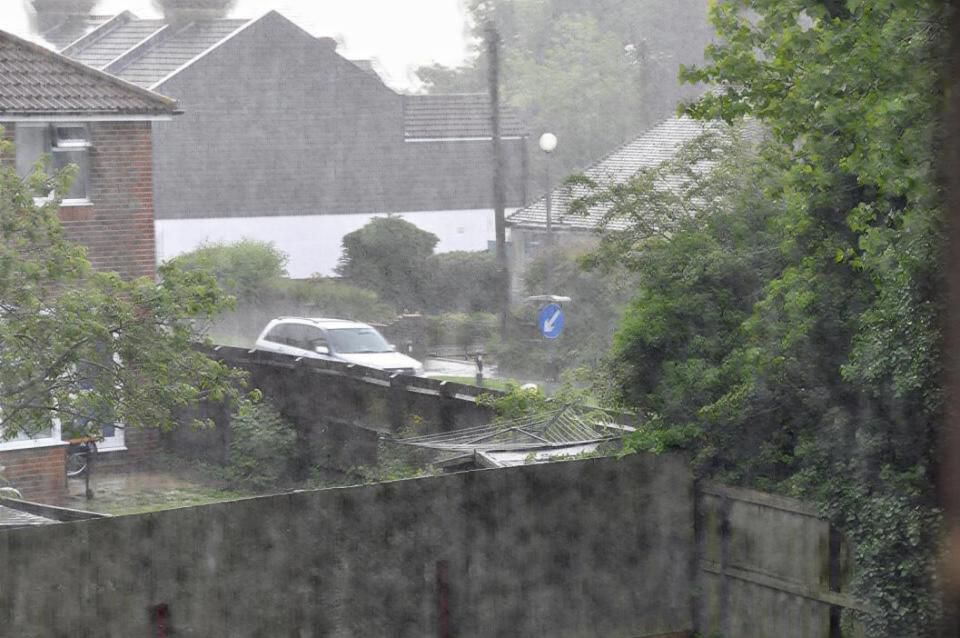}}
\subfigure[DID-MDN] {\includegraphics[height=0.7in, width=1.1in]{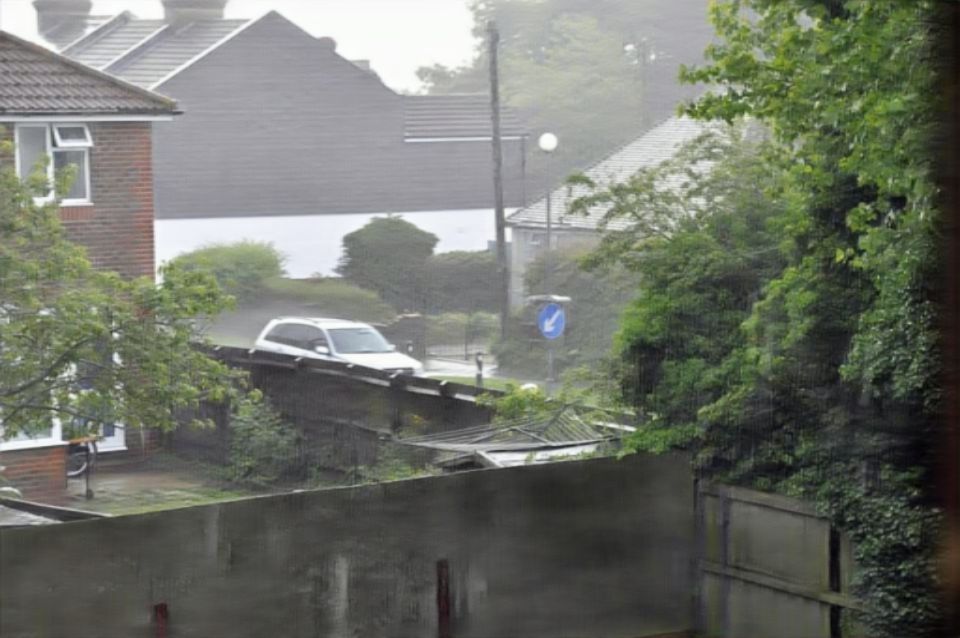}} 
 \subfigure[ResGuideNet]{\includegraphics[height=0.7in, width=1.1in]{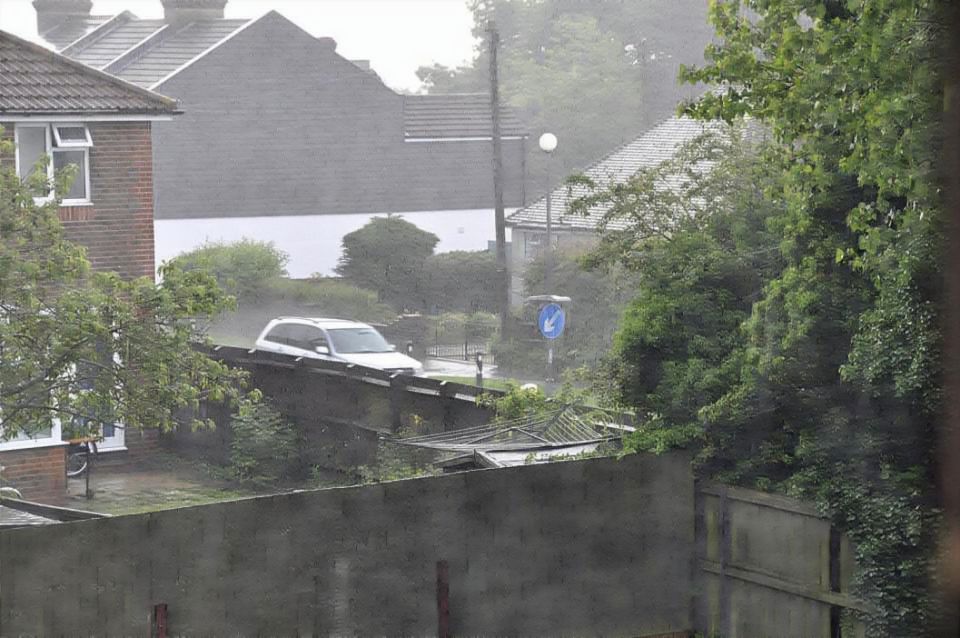}}

\caption{Five results on real-world rainy images with different rain magnitudes and shapes.}
\end{figure*}
\begin{table*}[]
\centering
\caption{Averaged SSIM and PSNR value on synthesized images with their parameter number. {\color{red}Red} indicates the best and {\color{blue}blue} indicates the second best performance. }
\label{my-label}
\begin{tabular}{|l|l|l|l|l|l|l|l|}
\hline
           & Rainy images& GMM\cite{Li} & DDN\cite{Fu}             & JORDER\cite{joder}        & DID-MDN\cite{DID-MDN}& ResGuideNet$_3$     & ResGuideNet  \\ \hline
Rain100H   & 13.56/0.379 & 15.05/0.425  & 21.92/0.764       &{\color{red}26.54}/{\color{blue}0.835} & 17.39/0.612         & 24.74/0.815       & {\color{blue}25.25}/{\color{red} 0.841}   \\ \hline
Rain100L   & 26.90/0.838 & 28.66/0.865  & 32.16/0.936       & {\color{red}36.63}/{\color{red}0.974}  & 25.70/0.858          & 32.82/0.960       & {\color{blue}33.16}/{\color{blue}0.963}  \\ \hline
Rain12     & 30.14/0.855 & 32.02/0.910  &  {\color{blue}31.78}/0.900& {\color{red}33.92}/{\color{red}0.953}  & 29.43/0.904    & 29.19/0.936    &29.45/{\color{blue}0.938}      \\ \hline
Parameters & \multicolumn{1}{c|}{-} & \multicolumn{1}{c|}{-}  & 57,369             & 369,792                   & $\approx$135,800       & {\color{red}19,404}& {\color{blue}37,065}    \\ \hline
\end{tabular}
\end{table*}
\subsection{Inter-block Ensemble}
\cite{ensemble} first well studied the idea of ensemble learning which combines predictors instead of selecting a single predictor, ensemble learning has also introduced  in neural networks to improve performance. \cite{boosting} arranged a committee of neural networks in a simple voting scheme, and the final output predictions is based on the averaged result. Recently, \cite{25} \cite{densenet} using deep neural networks to deel with several computer vision tasks also use the ensemble technique.

Motivating by ensemble idea, we integrate all intermediate reconstruction of each block to form the best reconstruction which is aggragated by concatenation. As is shown in the bottom-right of Figure 2, the final reconstruction is obtained from the fusion of all intermediate reconstructions by a 1$\times$1 convolution.
Note that, we only use the merged result in section 4 since it is convenient for comparison in other sections. We refer the output with merging operation as ResGuideNet while the output of block$_i$ as ResGuideNet$_i$.
We can observe an improved result from Table 1. The experiment is conducted on the test dataset of \cite{Fu}.

\subsection{The Proposed Architecture}
As discussed, the proposed ResGuideNet consists of repeated blocks. Each block includes several convolutional kernels and a global shortcut. The ResGuideNet propagates rain streak residual information from shallow blocks into deeper ones. The network architecture is shown in Figure 2. The final reconstruction is obtained by concanating all intermediate outputs and compressed them into the final rain-streak residual. $\ell_2$+SSIM supervision is applied to guide each blocks and the final merged output.

Our basic network structure can be expressed as:
\begin{equation}\label{eq1}
\begin{aligned}
&F_{1}\left( X \right) = \hat r_{1}, \hat y_{1} = X + \hat r_{1}  \\
&F_{2}\left(\hat y_{1} ; \hat r_{1} \right) = \hat r_{2}, \hat y_{2} = X + \hat r_{2} \\
&F_{3}\left(\hat y_{2}; \hat r_{2}  ,\hat r_{1} \right) = \hat r_{3}, \hat y_{3} = X + \hat r_{3}\\
&\cdots \cdots \\
&F_{5}\left(\hat y_{4}; \hat r_{4}  ,\hat r_{3},\hat r_{2},\hat r_{1} \right) = \hat r_{5}, \hat Y = X + \hat r_{5}\\
\end{aligned}
\end{equation}

where $F$ indicates different blocks that consists of several convolutional layers using Leaky Rectified Linear Units. $X$ and $Y$ indicate rainy and clean pairs. $\hat r$ indicates
negtive residual that is the output of each block. Block$_i$'s input is expressed as $\hat y_{i-1}$ . Note that the left side of the semicolon indicates input of each block while the right side indicates residual features to guide each block. It is shown that more guidance provided when the blocks go deeper. $\hat r_1$, $\hat r_2$, $\hat r_3$ $\cdots$ $\hat r_N$ all should be approximated to $Y-X$ in training stage as indicated in Equation 1, thus it is easier for deeper blocks to learn new rain streaks information with the guidance of rain streaks residual in shallow blocks.
\section{Experiments}
We compare our algorithm with several state-of-the-art deep and non-deep techniques on synthetic and real-world datasets.

\subsection{Implementation details}
We train and test the algorithm using TensorFlow for the Python environment on a NVIDIA GeForce GTX 1080 with 8GB GPU memory. We use the Xavier method to initialize the network parameter and RMSProp for parameter learning. We select the initial learning rate to be 0.001. We set the size of training batch to 16. 50000 iterations of training were required to train ResGuideNet. For all experiments we set the filter size to be $3\times 3$ except the merge convolution and each convolution layer has 16 feature maps.
\subsection{Dataset}
Since clean and rainy image pairs from real-world is hard to obtain, four synthetic datasets are aviable for comparison. \cite{joder} provide $Rain100H$ and $Rain100L$ that is synthesized with heavy and light rain, each of them contains 100 images for test. The third dataset called $Rain12$ collected by \cite{Li} which contains 12 syhthetic images. The last one is provided by \cite{Fu} constains 10K pairs of rainy/clean images with different orientations and magnitudes of rain streaks. For fair comparision, we train deep learning-based models and test them on synthetic datasets, one for $Rain100H$ and for $Rain100L$, the model trained on $Rain100L$ is used to test $Rain12$. During training stage, We randomly generate 0.8 million rainy/clean
patch pairs with size of 128$\times$128 in the training stage.

%
%

\subsection{Evaluation on Synthetic dataset}
We train and test all the methods with the same dataset \cite{joder} \cite{Li} except DID-MDN since the training code has not published. SSIM \cite{ssimloss} and PSNR are adopted to perform quantitative evaluations shown in Table 2. Our method has a comparable SSIM values with JORDER while outperforming other methods, which
is in consistent with the visual result. We can observe the intermediate result in the third block(ResGuideNet$_3$) even has a decent result compared with other methods in Figure 9.
However, our ResGuideNet contains far fewer parameters than others and can be sliced into a smaller
network to meet light rain condition with limited resources, potentially making ResGuideNet easily implemented in varying real-world applications.
\subsection{Evaluation on Real-world dataset}
In this section, we show that ResGuideNet trained on synthetic training data still works well on real-world application.
We implement other methods according to their optimal setting. Figure 10 show visual results on real-world rainy images.
Since no ground truth exists, we only show their qualitative result. As shown, ResGuideNet generate a less blurred result and have promising results on multiple kind of rain streaks.

\subsection{Running Time}
To illustrate the efficiency of implementation of ResGuideNet in practical application, we show the average running time of 100 test images in Table 3, all the test are conducted with a 500$\times$500 rainy image as input. The GMM is non-deep method that is run on CPUs according to the provided code, while other deep-based methods are tested on both CPU and GPU. All experiments are performed with the same environment described in implementation details. The GMM has the slowest running time since it has complicated inference at test time. Our method has a fast computational time on GPU compared with other methods. In a light rain condition, we can use the third blocks as final output for testing that has a even faster running time. This experiment shows ResGuideNet a promising practical value.
\begin{table*} [htbp]
 \caption{\label{table1} Running time of different methods.\\}
 \centering
 \begin{tabular}{lccccccl}
  \toprule
     & GMM & DDN & JORDER & DID-MDN & ResGuideNet$_3$ & ResGuideNet \\
  \midrule
  CPU         & 1990 & 1.51 & 295 & 4.20  & 1.26 & 3.15 \\
  GPU         & -     & 0.16 & 0.18 & 0.14 & 0.06 & 0.11  \\
  \bottomrule
 \end{tabular}
\end{table*}

\section{Extension}
\subsection{Generalization to other image processing tasks}
In this section, we show more evaluations for other general image processing tasks. We trained our ResGuideNet with the train and val set of berkeley segmentation dataset 500(BSD500) which contains 300 images in our training stage and we tested our model on the test set of BSD500 contains 100 images. We apply Gaussian noise with the standard deviation of 0.1 to both train and test datasets. The averaged SSIM on our test dataset is 0.927. We can see the reconstruction quality in Figure 11. This experiment demonstrates that ResGuideNet can generalize to similar image restoration problems.

\subsection{Pre-processing for high-level vision tasks}
Most exsting models for high-level tasks is trained with a well scenario, the performance will be degrated in rainy conditions since rain streaks block and blur the key structure of objects, Figure 12 show a case that under heavy rain condition, the pre-trained Faster R-CNN \cite{faster-rcnn} model trained on a well condition failed to capture some objects and produce a low recognition confidence. We incorporate
our ResGuideNet as a pre-process model for the Faster R-CNN, the detection performance has a great improvement over the naive Faster R-CNN input with a degraded image.
\begin{figure}
\subfigure[Noisy image] { {\includegraphics[width=1.6in]{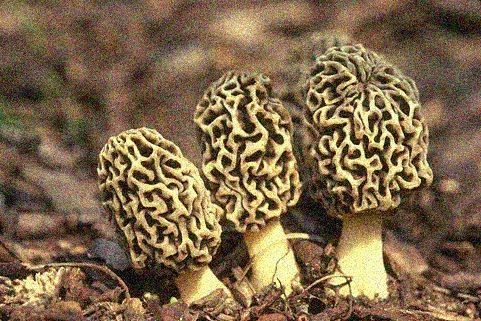}}}
\subfigure[Our result] { {\includegraphics[width=1.6in]{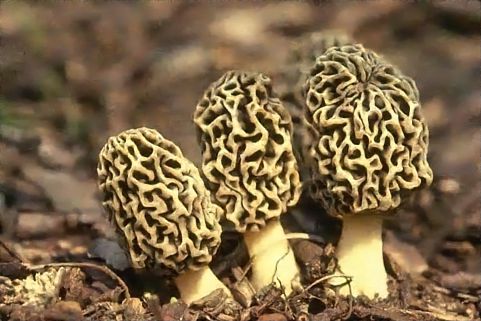}}}
\caption{An example of ResGuideNet applied to single-image denoising(SSIM = 0.964)}
\end{figure}
\begin{figure}
\subfigure[Rainy image] { {\includegraphics[height=2.0in, width=3.3in]{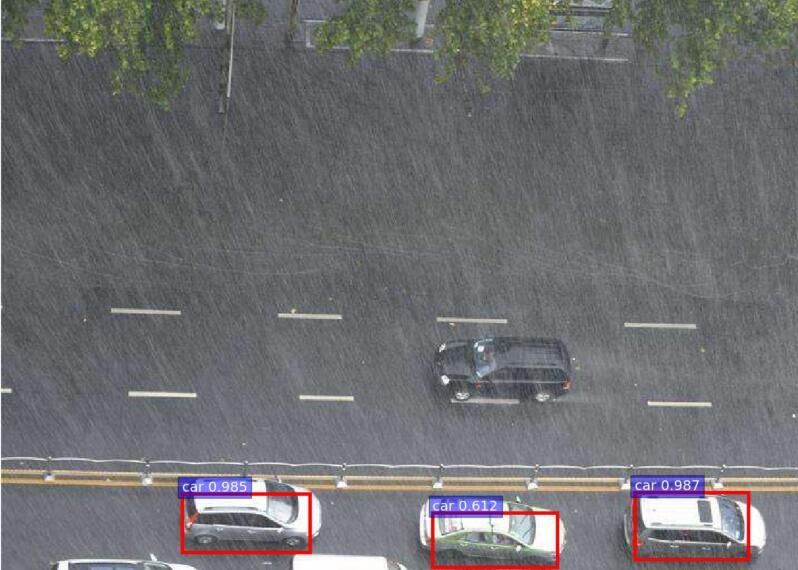}}}
\subfigure[Our result] { {\includegraphics[height=2.0in, width=3.3in]{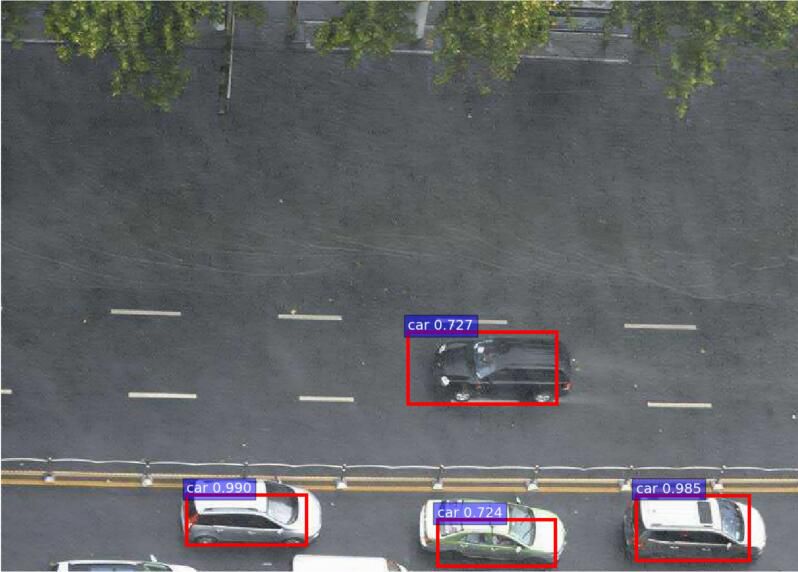}}}
\caption{An example of detection.}
\end{figure}
\section{Conclusion}
We presented the ResGuideNet, a novel convolutional network architecture for single image deraining which is easy to implement in a number of practical applications. We build our model with several derain sub-networks in a cascaded manner. By propagating negative residuals in shallow blocks to deeper ones, the deeper blocks effectively extract new information of negative rain streak residuals to get rain residual in a coarse to fine fashion. The final reconstruction take all intermediate outputs into account to leverage more informations across blocks which can be viewed as ensemble learning. With our proposed architecture, ResGuideNet has $37K$ and ResGuideNet$_3$ has less than $20K$ parameters while still achieving good performance. For different rain conditions and computational resources, we can detach ResGuideNet into a smaller size can still achieve decent reconstruction. Moreover, extensive experiments have shown that our ResGuideNet can generalize to other low-level tasks has potential value for high level vision problems.

\bibliographystyle{ACM-Reference-Format}
\bibliography{sample-bibliography}

\end{document}